%% file: root.tex
\documentclass[letterpaper, 10pt, conference]{ieeeconf}  % Comment this line out if you need a4paper

\IEEEoverridecommandlockouts                              % This command is only needed if 
\usepackage[utf8]{inputenc}
\usepackage[T1]{fontenc}
\usepackage{mathptmx} % assumes new font selection scheme installed
\usepackage{times} % assumes new font selection scheme installed
\usepackage{amsmath} % assumes amsmath package installed
\usepackage{amssymb}  % assumes amsmath package installed
\usepackage{algorithm}
\usepackage{algorithmic}
\floatname{algorithm}{\footnotesize Algorithm}
\usepackage{float}
\usepackage{stfloats}
\usepackage[maxfloats=256]{morefloats}
\usepackage{pgfplots}
\usepackage{array}
\usepackage{booktabs}
\usepackage{enumerate}
\usepackage{calc}
\usepackage{multirow}
\usepackage{tabularx}
\usepackage[caption=false]{subfig}
\usepackage[firstpage=true, placement=top]{background}
\backgroundsetup{contents={\parbox{\textwidth+1cm}{\copyright 2021 IEEE.  Personal use of this material is permitted.  Permission from IEEE must be obtained for all other uses, in any current or future media, including reprinting/republishing this material for advertising or promotional purposes, creating new collective works, for resale or redistribution to servers or lists, or reuse of any copyrighted component of this work in other works.}}, color=black, opacity=1, scale=0.9, vshift=-10pt}
\usepgfplotslibrary{groupplots}
\usepackage{tikz}
\usetikzlibrary{shapes,arrows,plotmarks,positioning, decorations.pathreplacing}
\pgfplotsset{compat=1.8}

\makeatletter
\let\NAT@parse\undefined
\makeatother
\usepackage{hyperref}

\title{\LARGE \bf
Human Initiated Grasp Space Exploration Algorithm for an Underactuated Robot Gripper Using Variational Autoencoder
}

\author{
\authorblockN{Clément Rolinat, Mathieu Grossard}
\authorblockA{Université Paris-Saclay, CEA, List,\\
F-91120, Palaiseau, France\\
\{clement.rolinat, mathieu.grossard\}@cea.fr}
\and
\authorblockN{Saifeddine Aloui, Christelle Godin}
\authorblockA{Université Grenoble Alpes, CEA, Leti,\\
F-38000, Grenoble, France\\
\{saifeddine.aloui, christelle.godin\}@cea.fr}
}

\begin{document}

\maketitle
\thispagestyle{empty}
\pagestyle{empty}

%%%%%%%%%%%%%%%%%%%%%%%%%%%%%%%%%%%%%%%%%%%%%%%%%%%%%%%%%%%%%%%%%%%%%%%%%%%%%%%%
\begin{abstract}

Grasp planning and most specifically the grasp space exploration is still an open issue in robotics. This article presents an efficient procedure for exploring the grasp space of a multifingered adaptive gripper for generating reliable grasps given a known object pose. This procedure relies on a limited dataset of manually specified expert grasps, and use a mixed analytic and data-driven approach based on the use of a grasp quality metric and variational autoencoders. The performances of this method are assessed by generating grasps in simulation for three different objects. On this grasp planning task, this method reaches a grasp success rate of 99.91\% on 7000 trials.

\end{abstract}


%%%%%%%%%%%%%%%%%%%%%%%%%%%%%%%%%%%%%%%%%%%%%%%%%%%%%%%%%%%%%%%%%%%%%%%%%%%%%%%%
\section{INTRODUCTION}

\input{introduction_short.tex}

\section{PROBLEM STATEMENT \& TOOL USED}
\label{sec:prob_state}

\input{prob_statement.tex}

\section{GENERAL WORKFLOW}
\label{sec:workflow}

\input{general_workflow.tex}

\section{WORKFLOW IMPLEMENTATION}
\label{sec:impl}

\input{workflow_impl.tex}

%\clearpage

\section{GRASP PLANNING TRIALS}
\label{sec:planning_test}

\input{planning_trial.tex}

%\addtolength{\textheight}{-8cm}   % This command serves to balance the column lengths

\section{CONCLUSION}

\input{conclusion.tex}

\bibliographystyle{IEEEtran}
\bibliography{references.bib}

\end{document}

%% file: introduction_short.tex
Grasping is fundamental in most of the industrial manufacturing processes such as pick-and-place, assembly or bin picking tasks. The grasp planning question is still an active research topic. It aims at finding a gripper configuration that allows to grasp an object reliably. This grasp configuration needs to be kinematically reachable and collision free with respect to the environment, and the produced grasp needs to be stable and robust to external perturbation. Finding such a grasp configuration requires to explore the grasp space, that is the subset of gripper configurations that effectively grasp the object. Thus, grasp planning is both object-dependent and hardware-dependent. Taking into account those constraints during the exploration is not trivial, as objects can have complex shapes, and gripper-arm combination can have complex kinematics. 

This is even more true for underactuated or compliant architectures, which are often chosen for grasping tasks \cite{townsend_barretthand_2000}. Indeed, such architecture allows to reduce the controller complexity by reducing the number of controlled degrees of freedom, while retaining sufficient kinematic abilities. Moreover, it tends toward producing robust grasps by their mechanical structure.

The grasp planner should be able to find in the high dimensional and highly constrained grasp space a configuration that fulfills a given criterion. There are two main ways to achieve this: analytic approaches and data-driven approaches \cite{sahbani_overview_2012}. Analytic approaches rely on an analytic description of the grasping problem \cite{berenson_grasp_2007}\cite{roa_grasp_2008}\cite{xue_grasp_2007}. Data-driven approaches depend on machine learning methods to predict grasps from object depth map or point cloud \cite{zhao_grasp_2020}\cite{pinto_supersizing_2015}\cite{depierre_jacquard:_2018}\cite{levine_learning_2018}\cite{mahler_dex-net_2017}\cite{Riedlinger_Model_2020}\cite{Mousavian_graspnet_2019}.

A shared issue is the grasp dataset creation, that is the grasp space exploration. A variety of high quality grasps need to be discovered by exploring the space of possible grasp configurations. There is two main approaches regarding this exploration \cite{xue_grasp_2007}: contact point approaches, and gripper configuration approaches. In the first case the grasp space exploration comes down to test various combination of contact point location on the object surface. However there is no guarantee that a given combination is kinematically admissible for a given gripper, and the inverse kinematics can even be intractable for underactuated or adaptive grippers. In the second case, the grasp space is explored by testing several gripper spatial configurations. This is more suited for underactuated gripper. Nevertheless, there is no assurance that a given gripper configuration is a priori able to grasp the object without realizing extensive simulation trials beforehand \cite{Riedlinger_Model_2020}\cite{Mousavian_graspnet_2019}.

To circumvent this dimensionality issue related to the huge size of the grasp space, numerous contact point approaches limit their search to fingertip contacts \cite{roa_grasp_2008}\cite{zhao_grasp_2020}, and gripper configuration approaches often use a bi-digital gripper and limit their search to planar grasps \cite{pinto_supersizing_2015}\cite{depierre_jacquard:_2018}\cite{levine_learning_2018}\cite{mahler_dex-net_2017}. For more complex grippers, as multifingered and adaptive ones, a human input is often required. For example, in Santina et al. \cite{santina_learning_2019}, authors identified a set of ten grasp primitives from human examples, and reduced the grasp space to those primitives only. In Choi et al. \cite{choi_learning_2018}, authors choose to limit the search space by discretizing it.

The contribution of this article is a procedure that allows to explore efficiently the grasp space of a multifingered and underactuated gripper. It relies on a limited set of object-dependent primitive gripper configurations, that are likely to grasp the object based on human experience, around which the exploration is focused. This allows to reduce the search space dimensionality, without restricting the kinematic potential of the gripper with arbitrary and strict hypothesis such as fingertip contacts or planar grasps. In this article, this procedure is applied in simulation on three different objects, and allows to successfully generate relevant grasps.

\hyperref[sec:prob_state]{Section~\ref*{sec:prob_state}} is dedicated to the problem statement and a presentation of the used framework. Then, in \autoref{sec:workflow} the general workflow will be explained. In \autoref{sec:impl} some implementation details will be given. Finally, the resulting grasp space exploration algorithm will be tested on grasp planning trials in \autoref{sec:planning_test}. To conclude, this work will be discussed and the planned future works will be presented.

%% file: prob_statement.tex
\subsection{Simulation Setup}

The gripper simulated in this work has three fingers, and is underactuated and adaptive. The compliance and underactuation allow it to naturally adapt itself to the object geometry, without the need to carefully control each joint, thus increasing the robustness of the grasp.

This gripper has two joints on each finger and one actuator per finger to control both joints. The second (distal) phalanx starts moving when the applied effort on the finger is above a given force threshold. A fourth actuator allows a coupled and symmetrical abduction-adduction (spread) motion of two fingers.

This gripper is mounted as end effector of a six degrees of freedom industrial robot arm.

The simulation setup described above is implemented with Gazebo simulator \cite{koenig_design_2004}. A picture of this simulated setup is displayed in \autoref{fig:gripper_info}.

\subsection{Problem Statement}
\label{sec:statement}

An object is placed on a table in the workspace of the considered robotic setup. It is assumed that the object is known, as well as the pose in the scene of its associated frame $F_{obj}$. Knowing the pose is not a strong hypothesis, as there exists methods to extract pose information of known objects from a point cloud, for example \cite{drost_model_2010}.

The goal of the grasp space exploration is to find grasp configurations with a high quality. The metric used to assess grasp quality will be described in \autoref{sec:metric}. In this work, a grasp configuration is a gripper configuration that is able to grasp the object without colliding with the table. 

\begin{figure}[ht!]
\centering
\subfloat[Pose of the frame $F_{grip}$ used to locate the gripper relatively to the object frame $F_{obj}$. \label{subfig:gripper_frame}]{\includegraphics[width=0.616\linewidth, keepaspectratio]{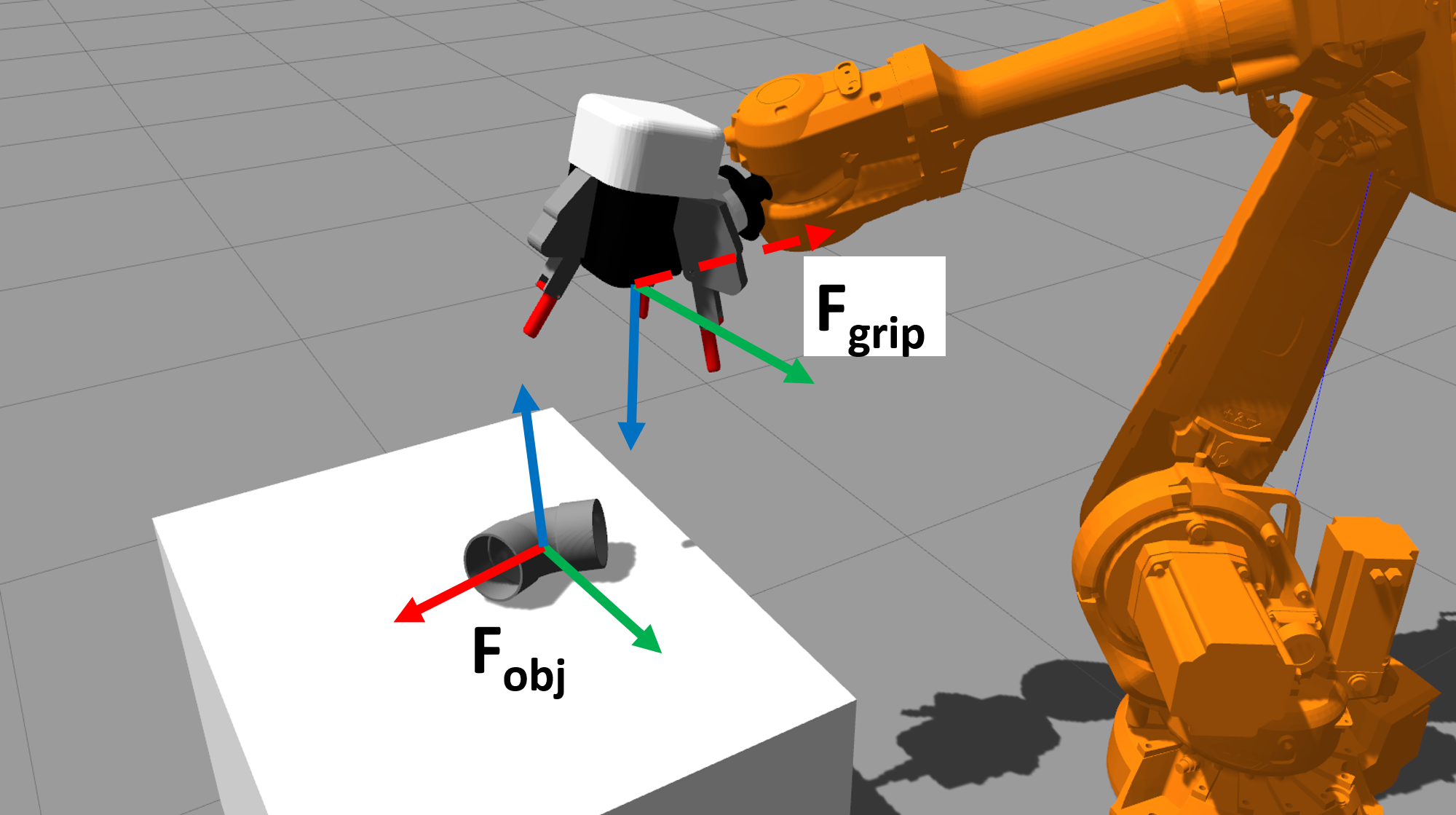}}
\subfloat[Spread angle $\theta$. \label{subfig:gripper_spread}]{\includegraphics[width=0.335\linewidth, keepaspectratio]{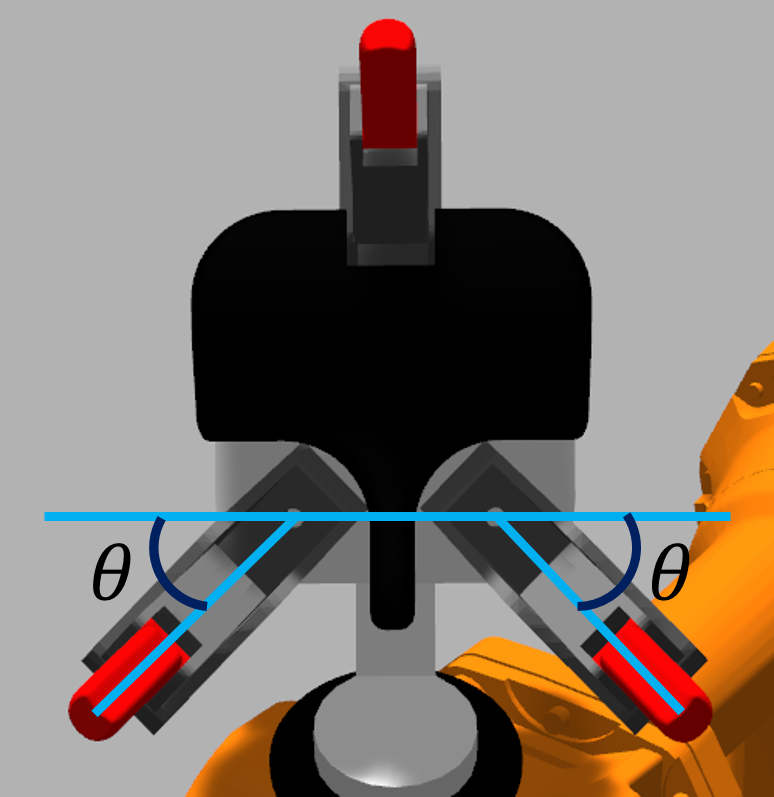}}
\caption{Gripper frame and spread angle.}
\label{fig:gripper_info}
\end{figure}

A gripper configuration is defined as follows by eight parameters:
\begin{itemize}
\item the pose of the gripper frame $F_{grip}$,
\[
(x, y, z, q_x, q_y, q_z, q_w) \in \mathbb{R}^{3} \times SO(3) = SE(3)
\]
with the orientation expressed in quaternion convention;
\item the abduction-adduction motion, or spread angle $\theta$, as shown on \autoref{fig:gripper_info}. 
\end{itemize}

The dimensionality of this configuration space is high, but this allows to fully leverage the grasping ability and kinematic potential of the gripper. Thus, the grasp space is a subset of this configuration space, with an additional constraint that every gripper configuration is able to grasp the object without colliding with the table.

To locate the gripper, a dedicated frame $F_{grip}$ situated between the fingers in front of the palm is used. This frame is displayed on \autoref{fig:gripper_info}. Gripper poses are expressed relatively to the object frame $F_{obj}$, in order to be invariant to object poses.

\subsection{Analytical Grasp Quality Metric}
\label{sec:metric}

A grasp has several properties which can define its quality \cite{roa_grasp_2015}. One of them is the \emph{force-closure} property, that is the ability to resist external disturbances in any direction. See Murray et al. \cite{murray_mathematical_1994} and Prattichizzo et al. \cite{prattichizzo_grasping_2008} for more in depth mathematical description. In particular, it relies on the concept of grasp map $G$, a matrix that stores geometrical information about the grasp.

Several metrics have been developed from the computation of the grasp map. Some of them are considering algebraic properties of the matrix, such as the full rank of the $G$ matrix, and can be used as a proxy for force-closure. In this paper, the \emph{minimum singular value of G}, $Q_{MSV}$, has been chosen. It is worth noting that the proposed grasp space exploration method could work with an other grasp quality metric. With $\sigma(G)$ the vector of singular values of $G$, $Q_{MSV}$ can be simply expressed as follows:

\[
Q_{MSV} = min \left(\sigma \left(G \right)\right)
\]

$Q_{MSV}$ greater than 0 is a necessary condition for force-closure. The greater $Q_{MSV}$ is, the farther the grasp is from a numerical singularity. However, this is not a sufficient condition. In the general case, it is difficult to assess if a grasp is force-closure because it comes down to an optimization problem with non-linear constraints.

\subsection{Variational Autoencoders}

A variational autoencoder (VAE) allows to generate consistent data from its latent space more reliably than a classic autoencoder, which is designed to learn a compressed representation of data.

In a VAE, among other features, a supplementary term is added in the loss function: the Kullback-Leibler (KL) divergence \cite{kingma_auto_2014}. This term helps the data to be represented as a normal distribution in its latent space, and thus regularizing it.

%% file: general_workflow.tex
\begin{figure*}
\centering
\smallskip
\input{scheme_general_workflow.tex}
\caption{Scheme of the presented workflow.}
\label{fig:scheme_workflow}
\end{figure*}
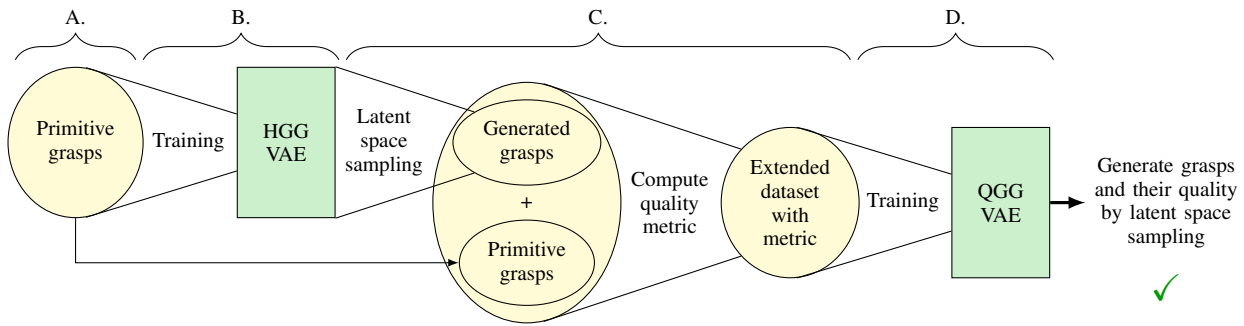

The idea presented in this paper is to take advantage of the human ability to find promising grasp configurations. Indeed, it is easy for a human to find gripper configurations that are likely to grasp a given object, that is configurations belonging to the grasp space. However, those primitive grasps do not necessarily have a high quality. Indeed, it is difficult for a human to assess a priori the relative and absolute quality of grasp configurations. Here, it is the role of the space exploration to focus around those primitive grasps to constitute a collection of grasps with various quality, and discover grasps with higher quality than the primitive ones if such grasps exist.

The general workflow used to achieve this is described below, and is decomposed as follows:
\begin{samepage}
\begin{enumerate}[A.]
\item the constitution of a primitive grasp dataset
\item the training of a Human-initiated Grasp Generator Variational Autoencoder (HGG)
\item a dataset extension \& grasp quality estimation phase
\item the training of a Quality-oriented Grasp Generator Variational Autoencoder (QGG)
\end{enumerate}
\end{samepage}

This procedure is summarized on \autoref{fig:scheme_workflow}.

\subsection{Primitive Grasp Dataset}

To leverage the human ability to find gripper configurations belonging to the grasp space, an object-dependent primitive grasp dataset is built. Concretely, a primitive grasp is a handcrafted gripper configuration, with its pose and spread angle human-chosen so that it is collision free and likely to grasp the object. The spread angle (shown in \autoref{fig:gripper_info}) is chosen between four discrete values corresponding to main gripper internal layouts: $\theta = 0$, $\theta = \pi/6$, $\theta = \pi/4$, and $\theta = \pi/2$.

The dataset stores the  eight parameters describing each primitive grasp along with the four parameters of the tabletop plane Cartesian equation in object frame. Indeed, many objects have different possible stable positions on the table. This is a critical information to avoid collisions with it. Some grasps may collide with the table in a given stable position, while being suitable for an other stable position. Expressing the grasp configuration in the object frame is still useful as it allows an invariance to a position change and to a rotation around a vertical axis.

\subsection{Human-initiated Grasp Generator Variational Autoencoder (HGG)}

The goal of the Human-initiated Grasp Generator VAE (HGG) is to extract the correlations existing between the parameters of different grasp primitives. Such correlations exist because primitive grasps are in the grasp space, and the grasp space is a subset of the gripper configuration space. A VAE is able to use those correlations to build an efficient mapping of the grasp space in its latent space, with fewer parameters than the initial gripper configuration space. This efficient dimension reduction allows to generate grasps sufficiently close to primitive grasps to remain pertinent, while exploring the configuration space around it, by simply sampling in the VAE latent space. In this case, a mono-dimensional latent space has been chosen, because it allows the strongest compression. The main drawback is that if the true dimensionality of the grasp space is greater than one, there will be information loss during compression. The effect of a latent space of higher dimension on the grasp space exploration needs to be investigated in future work.

The HGG is trained on the primitive grasp dataset. Its inputs and outputs are summarized on \autoref{fig:in_out}.

\begin{figure}[ht]
\centering
\input{input_output_table}
\caption{HGG inputs and outputs data. This input-output architecture is similar to Conditional VAE architecture introduced in \cite{sohn_learning_2015}. The generated grasp configuration is conditioned by the tabletop plane equation.}
\label{fig:in_out}
\end{figure}

For the gradient descent during the training, a Mean Square Error (MSE) is computed for each gripper parameter.

Each of these errors is averaged on each batch. Then, the global loss for each batch is computed as the sum of these averaged errors together with the KL-divergence loss.

\subsection{Dataset Extension \& Grasp Quality Estimation}

Sampling in the latent space of the HGG allows to explore the grasp space. This sampling is more efficient than a sampling around expert grasp in the gripper configuration space. Indeed, the HGG takes into account the correlations existing between the parameters of the grasp configurations, and thus maps the grasp space.

Each sampled configuration is then tested in simulation along with each primitive grasp to check its success.

A configuration is successful if the following conditions are met:
\begin{itemize}
\item it does not collide with the table
\item it successfully lifts the object from the table
\item its $Q_{MSV}$ is greater than 0.
\end{itemize}

For each successful configuration, the computed $Q_{MSV}$ quality value is registered. For failed configurations, a null value is registered as quality value. 

This allows to extend the primitive dataset by exploring extensively the grasp space. Thus, a collection of grasps with various quality values can be constituted, and if better grasps than the primitive ones exist, they can be discovered.

\subsection{Quality-oriented Grasp Generator Variational Autoencoder (QGG)}

The goal of the Quality-oriented Grasp Generator VAE (QGG) is to reliably generate grasps with their corresponding grasp quality.

The QGG is trained on the extended set formed by merging the primitive grasp set with the generated grasp set (both successful and failed). Learning failed grasps together with successful ones reduces the risk of predicting a high quality for a failing grasp when some failing configurations are close to successful ones. The inputs-outputs are the same as for the HGG with the grasp quality added as a supplementary output. This way, the QGG decoder learns to predict the grasp quality while reconstructing the other grasp configuration parameters. Moreover, the dataset extension allows to represent more accurately and more reliably the grasp space.

The latent space dimension and the loss function are the same as the ones used for the HGG. Regarding the grasp quality, a MSE is computed and added to the loss.

The QGG can be used to generate high quality grasps by sampling in its latent space and selecting only grasps having their quality above a given threshold.

%% file: scheme_general_workflow.tex
%\documentclass{standalone}
%\usepackage{mathptmx}
%\usepackage{times}
%\usepackage{amsmath}
%\usepackage{amssymb}
%\usepackage{tikz}
%\usetikzlibrary{shapes,arrows,plotmarks,positioning, decorations.pathreplacing}
%\begin{document}

\begin{tikzpicture}[every node/.append style={font=\footnotesize}]
\tikzstyle{dataset}=[draw, ellipse, fill=yellow!20]
\tikzstyle{vae}=[draw, fill=black!30!green!20]

\node[dataset, align=center, minimum height=2.cm](prim_grasp) {Primitive \\ grasps};
\node[vae, minimum width=1.3cm, minimum height=2.cm](HGGVAE_rect) at (2.8, 0) {};
\node[align=center, minimum width=1.3cm](HGGVAE) at (2.8, 0) {HGG \\ VAE};
\draw[] (prim_grasp.82) -- (HGGVAE.north west);
\draw[] (prim_grasp.-82) -- (HGGVAE.south west);
\node[] at(1.5, 0) {Training};
\node[dataset, minimum height=3.2cm, minimum width=2.5cm](grasp_ext) at (6, -0.8) {};
\node[dataset, align=center, minimum height=1cm](gen_grasp) at (6, 0) {Generated \\ grasps};
\draw[] (HGGVAE_rect.north east) -- (gen_grasp.north west);
\draw[] (HGGVAE_rect.south east) -- (gen_grasp.south west);
\node[align=center] at (4.1, 0) {Latent \\ space \\ sampling};
\node[dataset, align=center, minimum height=1cm](prim_grasp_met) at (6, -1.6) {Primitive \\ grasps};
\node[] at (6, -0.8) {+};
\draw[->, >=latex] (prim_grasp) |- (prim_grasp_met);
\node[dataset, align=center, minimum height=2.cm](grasp_met) at (9.5, -0.8) {Extended \\ dataset \\ with \\ metric};
\draw[] (grasp_ext.82) -- (grasp_met.north west);
\draw[] (grasp_ext.-82) -- (grasp_met.south west);
\node[align=center] at(7.9, -0.8) {Compute \\ quality \\ metric};
%\draw[->, >=latex, very thick] (grasp_ext.east) -- (grasp_met.west) node[midway, above, align=center] {compute \\ quality \\ metric};
\node[vae, minimum width=1.3cm, minimum height=2.cm](QGGVAE_rect) at (12.3, -0.8) {};
\node[align=center, minimum width=1.3cm](QGGVAE) at (12.3, -0.8) {QGG \\ VAE};
\draw[] (grasp_met.85) -- (QGGVAE.north west);
\draw[] (grasp_met.-85) -- (QGGVAE.south west);
\node[] at (11, -0.8) {Training};
\node[align=center](final) at (14.5, -0.8) {Generate  grasps \\ and their quality \\ by latent space \\ sampling};
\node[below = 0.1cm of final] {\textcolor{black!40!green}{\Large\checkmark}};
\draw[->, >=latex, very thick](QGGVAE.east) -- (final);
\draw[decoration={brace, amplitude=10pt},decorate]
  (-0.8,1.1) -- node[above=10pt] {A.} (0.8,1.1);
\draw[decoration={brace, amplitude=10pt},decorate]
  (0.9,1.1) -- node[above=10pt] {B.} (3.5,1.1);
\draw[decoration={brace, amplitude=10pt},decorate]
  (3.6,1.1) -- node[above=10pt] {C.} (10.3,1.1);
\draw[decoration={brace, amplitude=10pt},decorate]
  (10.4,1.1) -- node[above=10pt] {D.} (13,1.1);
\end{tikzpicture}

%\end{document}

%% file: input_output_table.tex
\footnotesize
\begin{tabular}{| *{12}{>{\centering\arraybackslash}m{0.283cm} |}}
\hline
\multicolumn{12}{|c|}{inputs} \\
\hline
\multicolumn{8}{|>{\centering}m{2.264cm + 14\tabcolsep + 2.8pt}|}{gripper configuration in frame $F_{obj}$} & \multicolumn{4}{>{\centering}m{1.132cm + 6\tabcolsep + 1.2pt}|}{tabletop plane Cartesian equation in frame $F_{obj}$} \\
\hline
%\multicolumn{3}{|c|}{position} & \multicolumn{4}{c|}{orientation} &  & \multicolumn{4}{c|}{equation coefficients} \\
%\hline
$x$ & $y$ & $z$ & $q_x$ & $q_y$ & $q_z$ & $q_w$ & $\theta$ & $a$ & $b$ & $c$ & $d$ \\
\hline
\multicolumn{8}{|c|}{outputs} & \multicolumn{4}{c}{} \\
\cline{1-8}
\end{tabular}

%% file: workflow_impl.tex
The workflow described above is performed on three different objects. Implementation details will be given in this section.

\subsection{Objects \& Primitive Grasps}

\begin{figure}[htb]
\setlength{\tabcolsep}{0pt}
\renewcommand{\arraystretch}{0}
\centering
%\medskip
\begin{tabular}{c c c}
\includegraphics[width=0.24\linewidth, keepaspectratio]{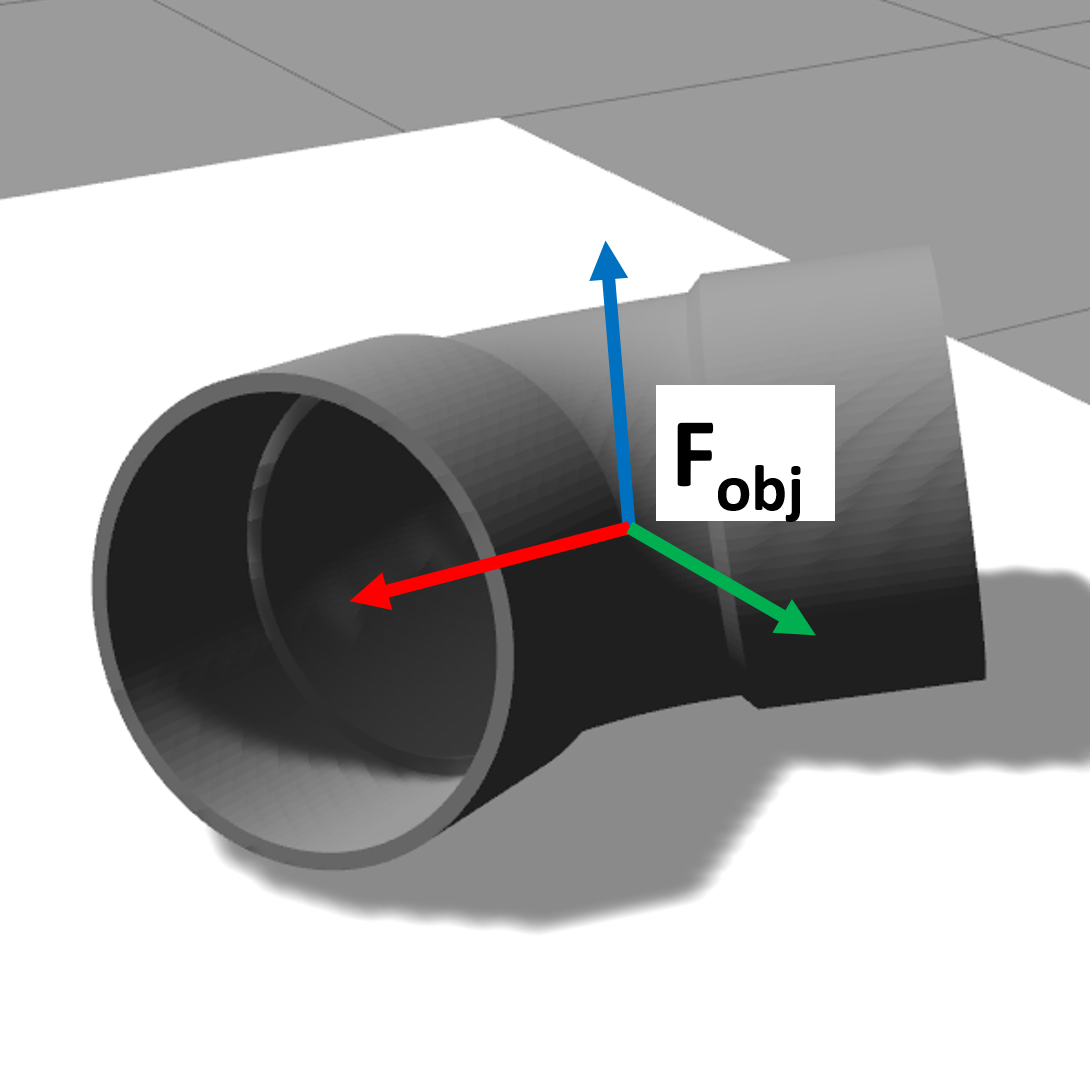} &
\includegraphics[width=0.24\linewidth, keepaspectratio]{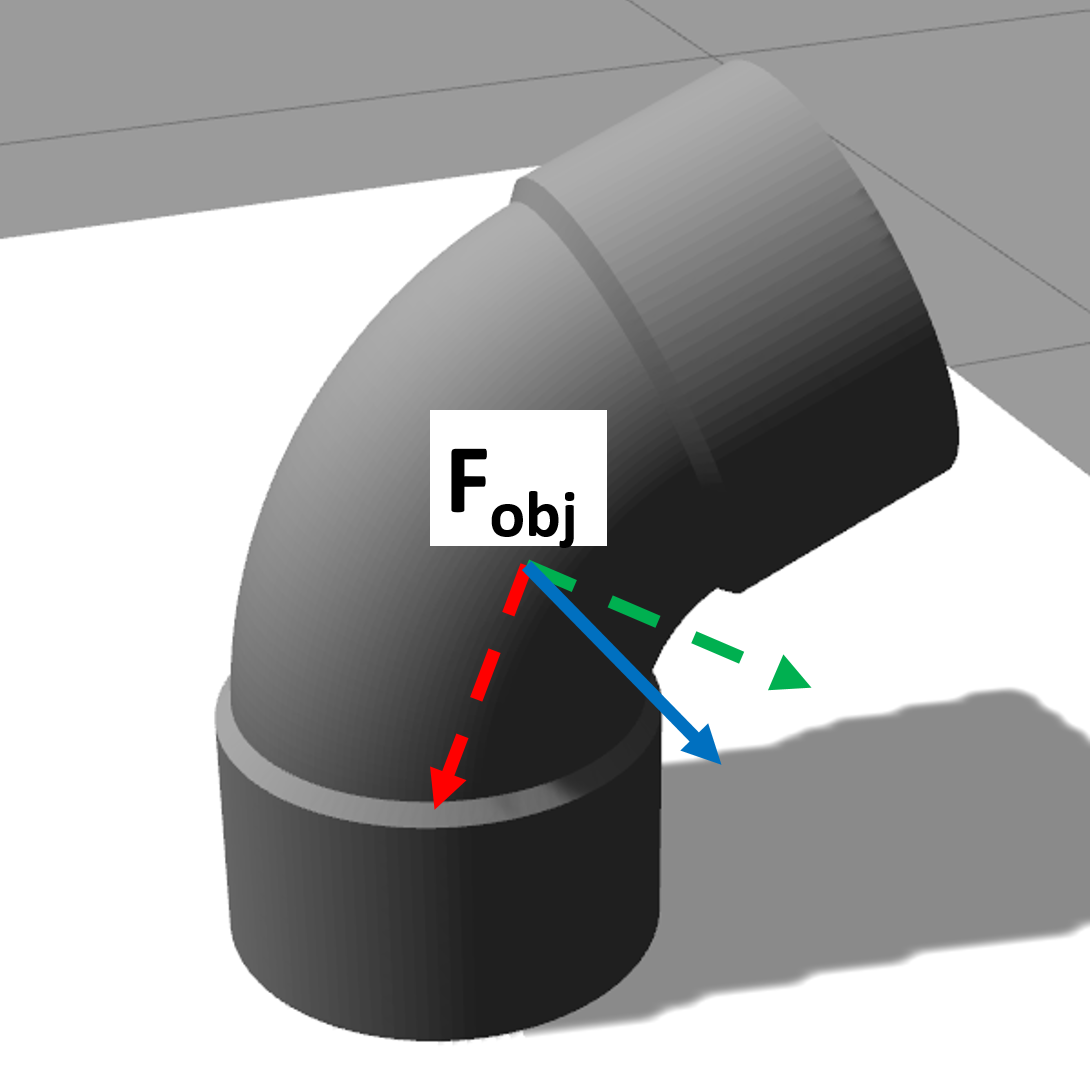} &
\\
\includegraphics[width=0.24\linewidth, keepaspectratio]{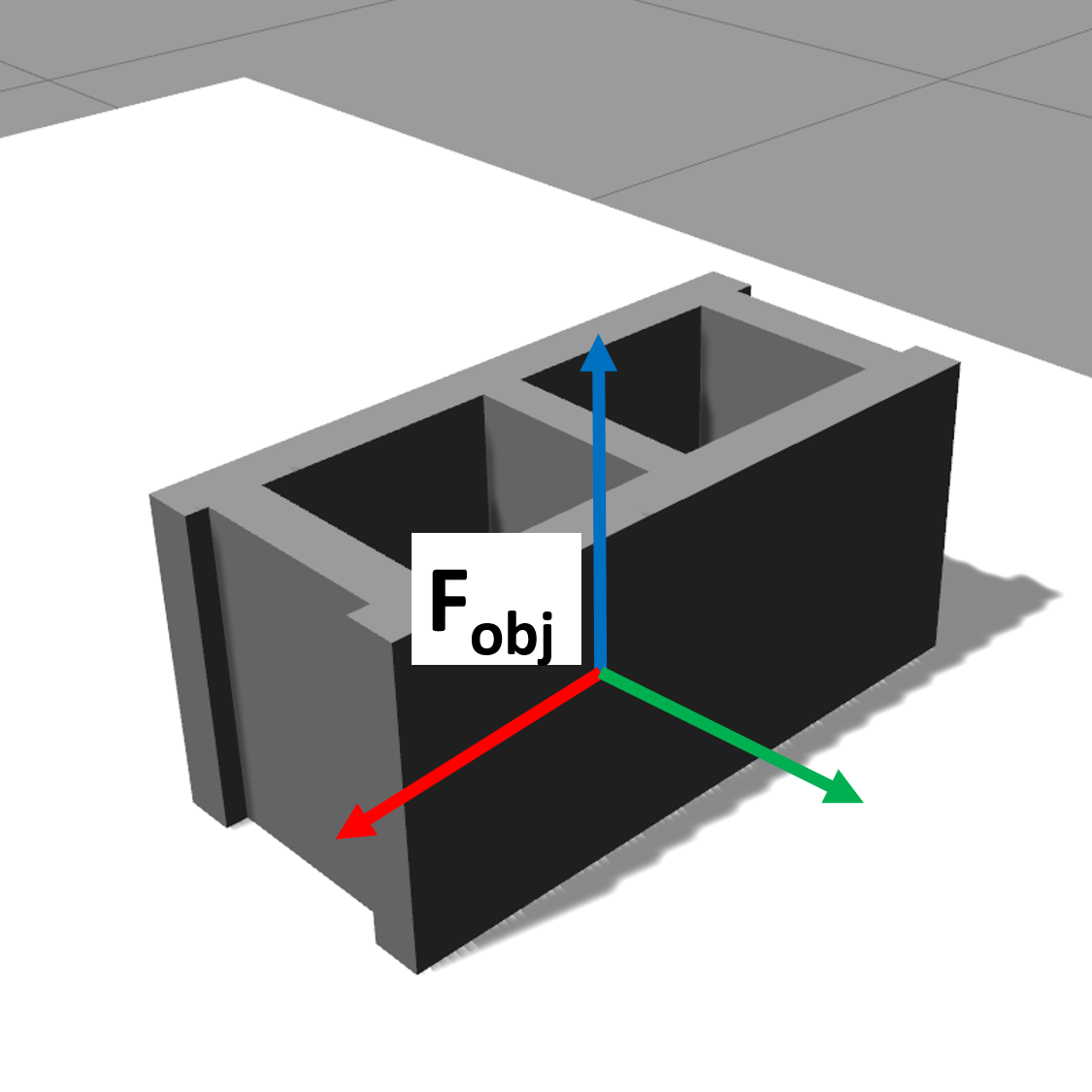} &
\includegraphics[width=0.24\linewidth, keepaspectratio]{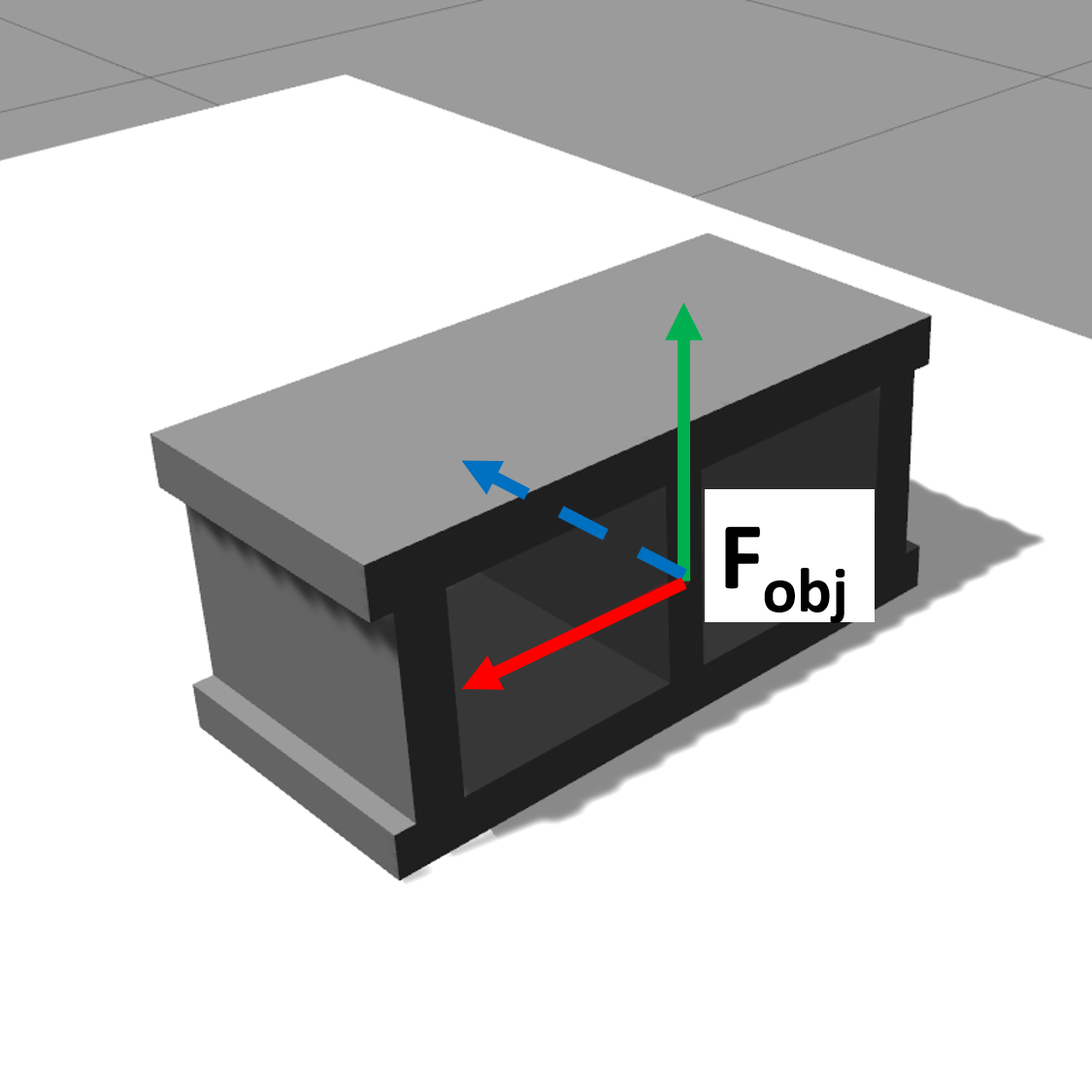} &
\includegraphics[width=0.24\linewidth, keepaspectratio]{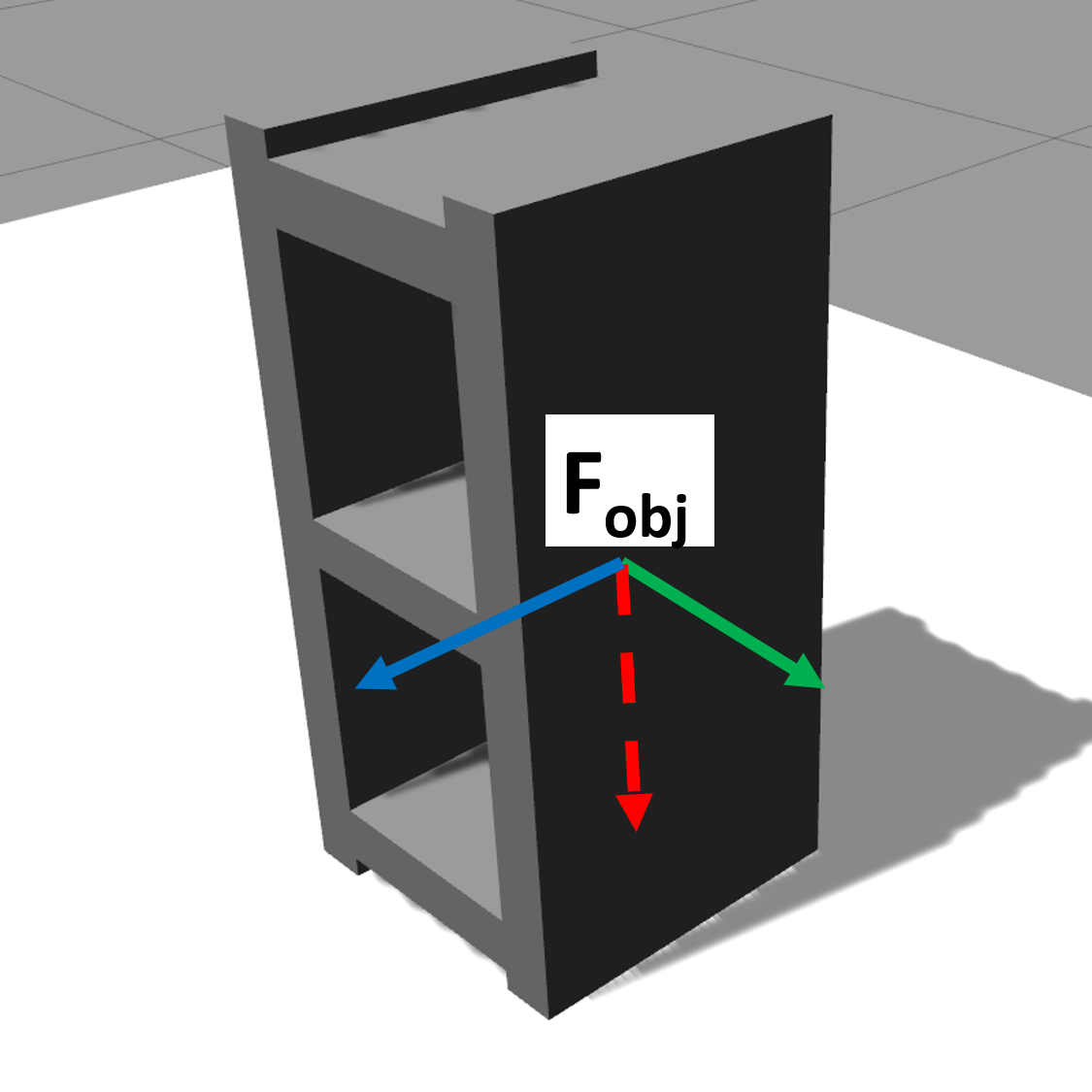}
\\
\includegraphics[width=0.24\linewidth, keepaspectratio]{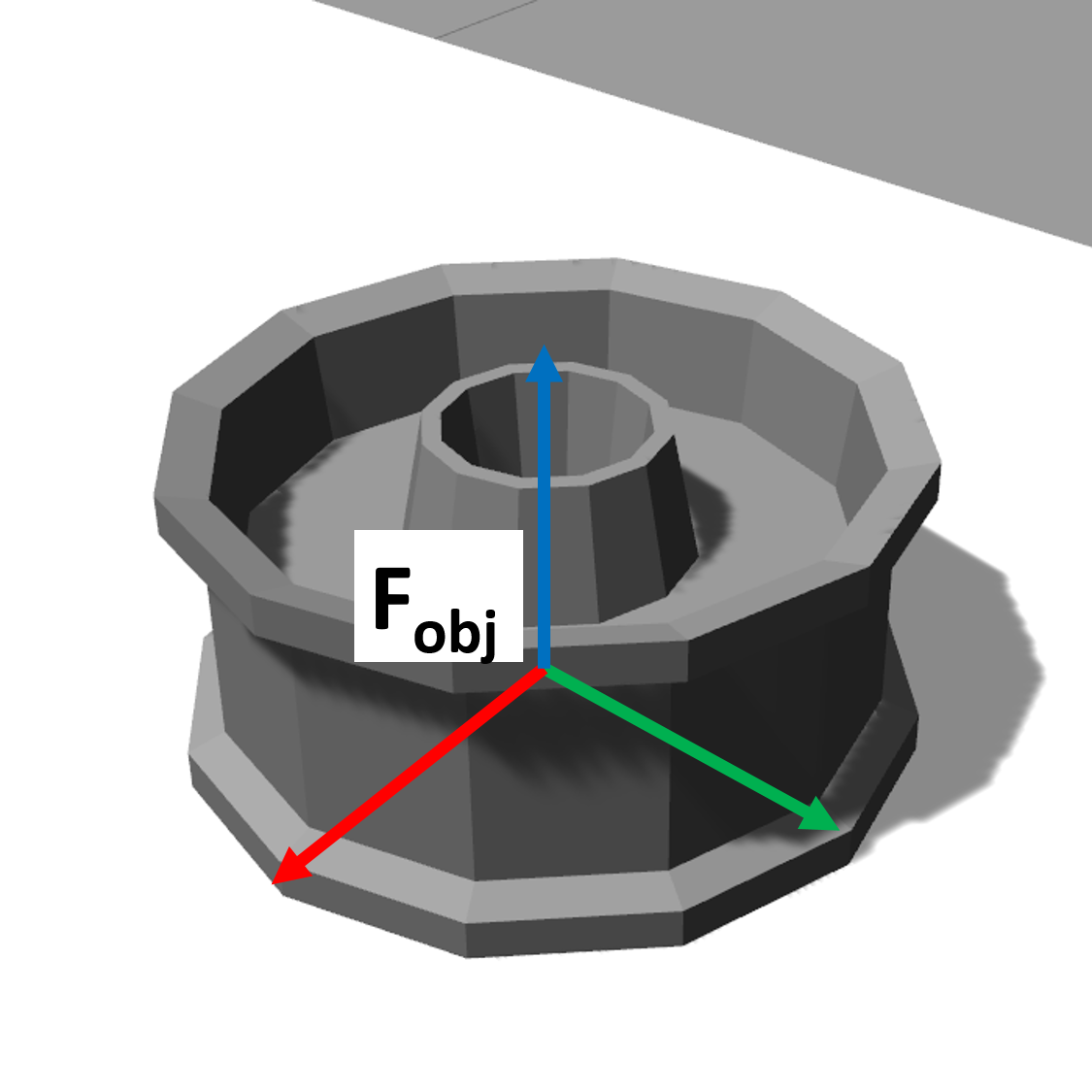} &
\includegraphics[width=0.24\linewidth, keepaspectratio]{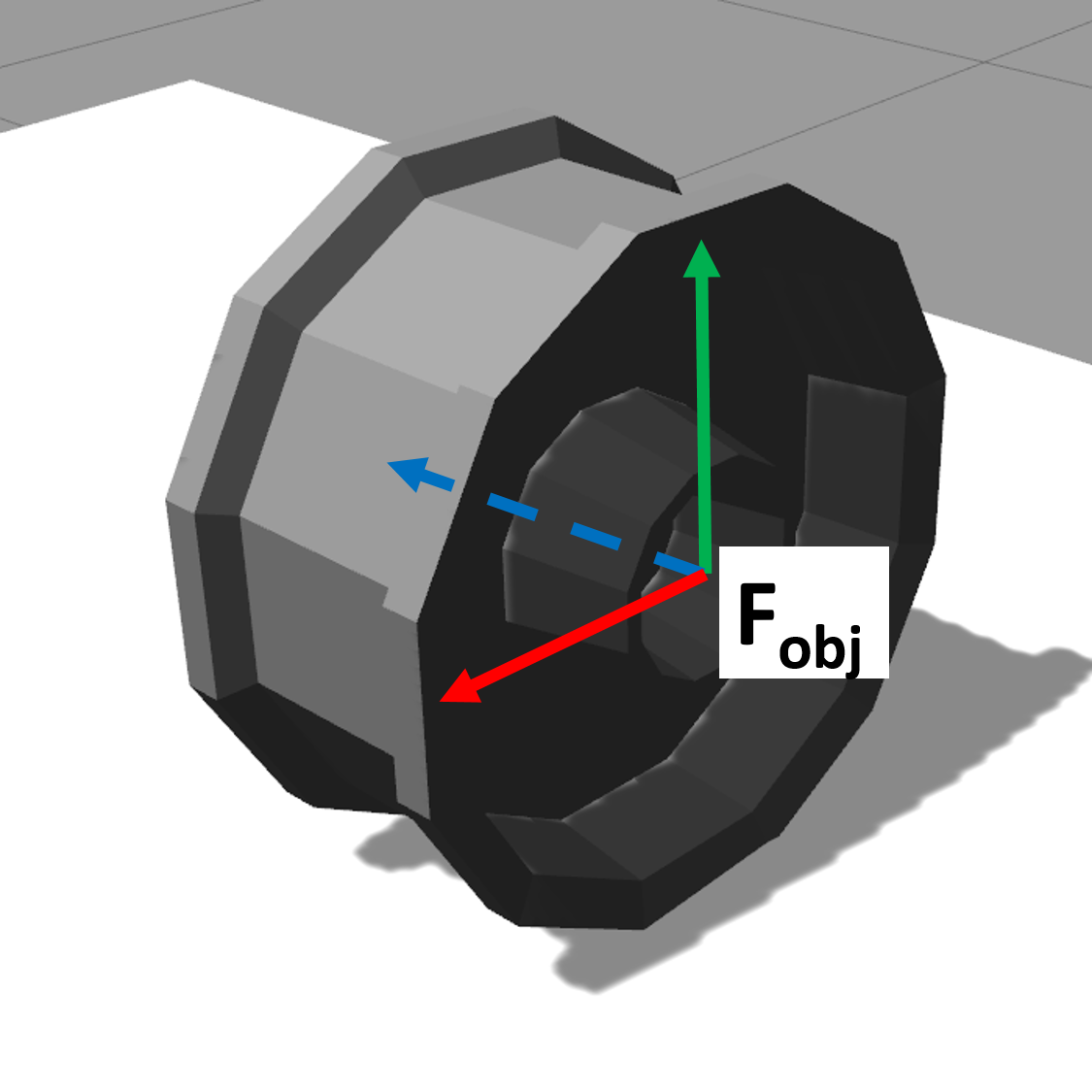} &
\end{tabular}
\caption{The chosen objects and their frame $F_{obj}$ in their different stable positions: bent pipe (first row), cinder block (second row), and pulley (third row)}
\label{fig:object_stable}
\end{figure}

The chosen objects are:
\begin{samepage}
\begin{itemize}
 \item a connector bent pipe
 \item a pulley
 \item a small cinder block
\end{itemize}
\end{samepage}

Their CAD model used in the simulation in their different stable positions are visible on \autoref{fig:object_stable}. Those object were chosen for their relative complexity and diversity in term of shapes.

\begin{figure}[H]
\setlength{\tabcolsep}{0pt}
\renewcommand{\arraystretch}{0}
\centering
\begin{tabular}{c c c c c}
\includegraphics[width=0.19\linewidth, keepaspectratio]{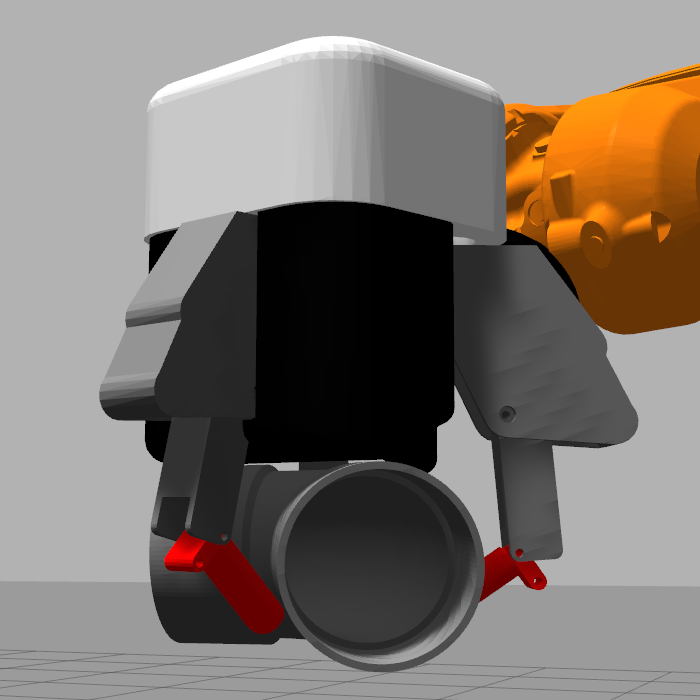} & \includegraphics[width=0.19\linewidth, keepaspectratio]{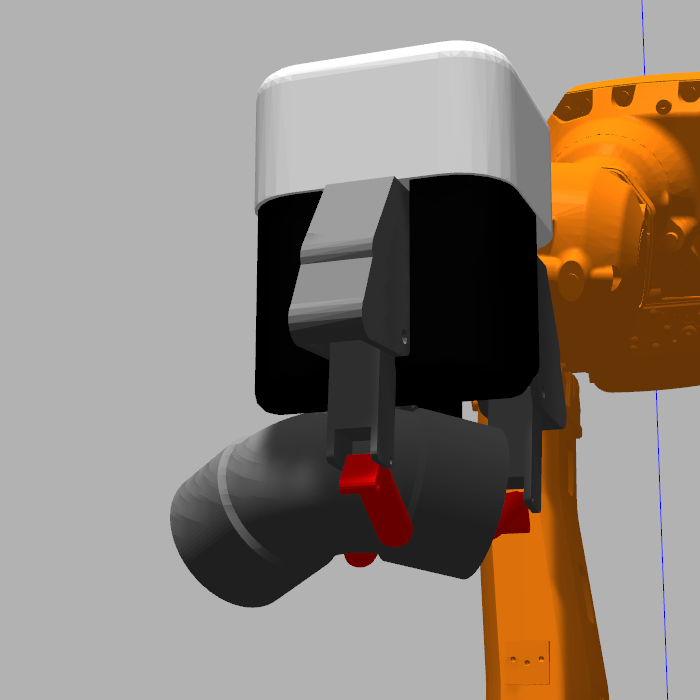} & \includegraphics[width=0.19\linewidth, keepaspectratio]{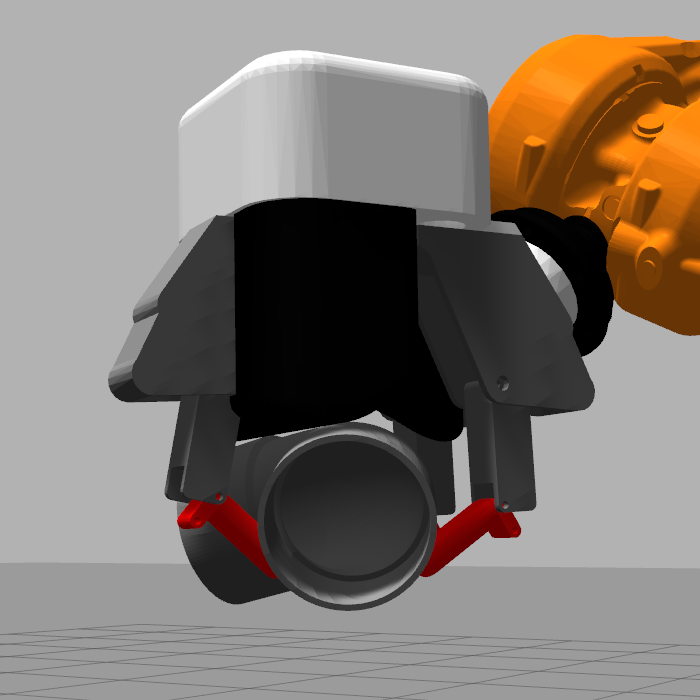} &
\includegraphics[width=0.19\linewidth, keepaspectratio]{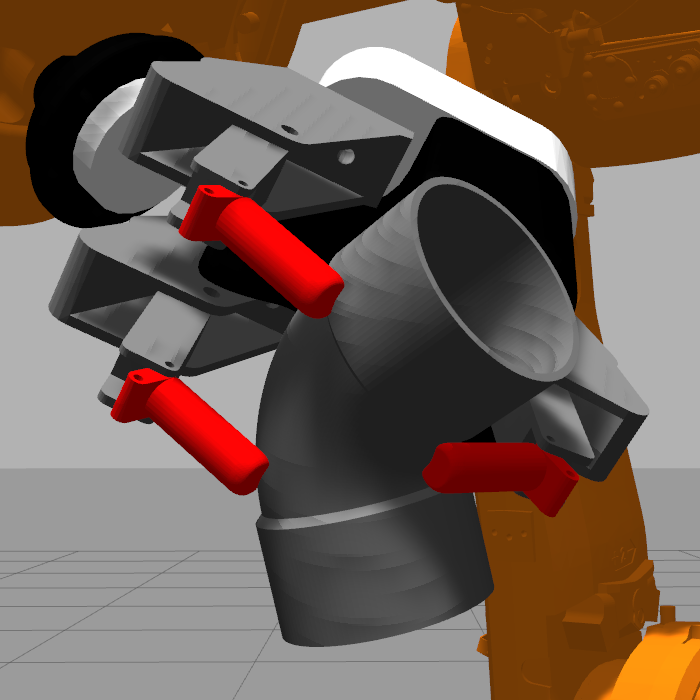} &
\includegraphics[width=0.19\linewidth, keepaspectratio]{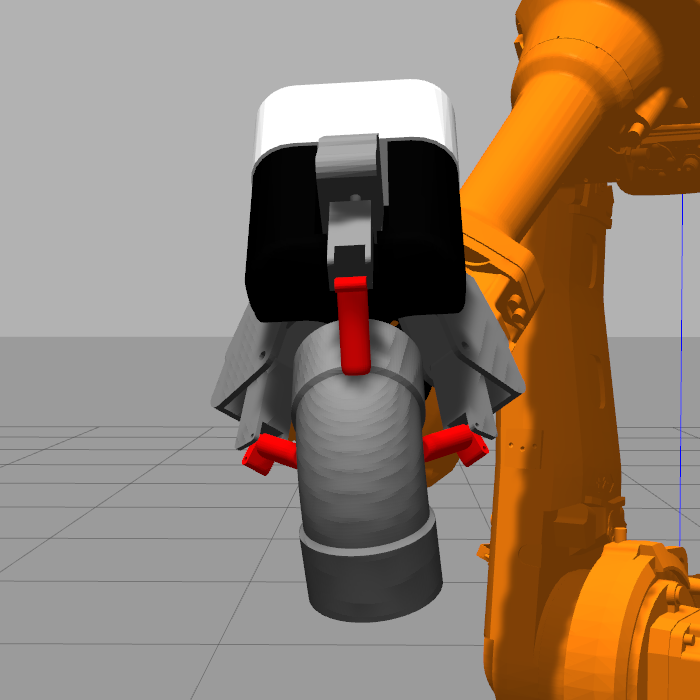}
\\
\includegraphics[width=0.19\linewidth, keepaspectratio]{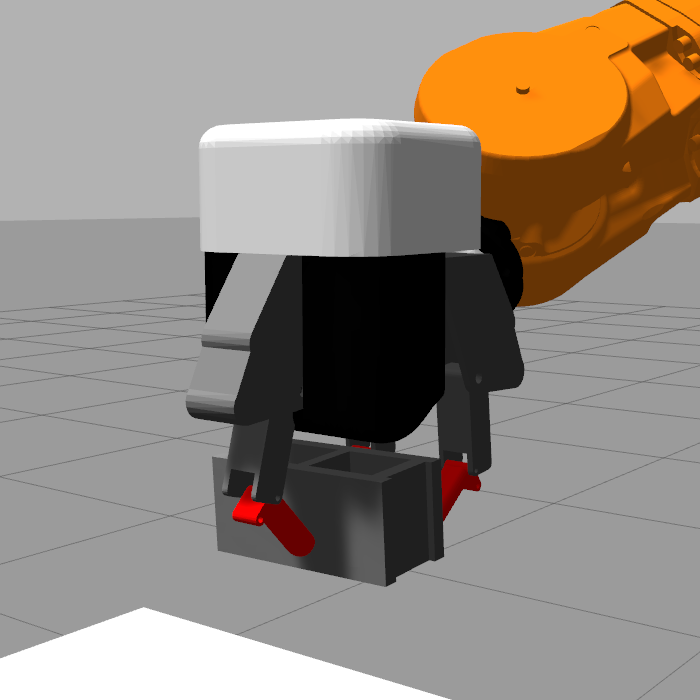} &
\includegraphics[width=0.19\linewidth, keepaspectratio]{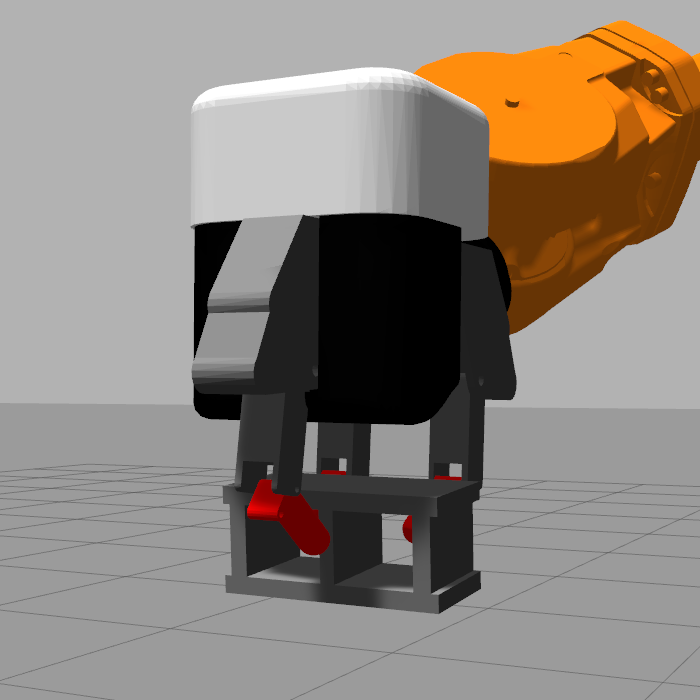} &
\includegraphics[width=0.19\linewidth, keepaspectratio]{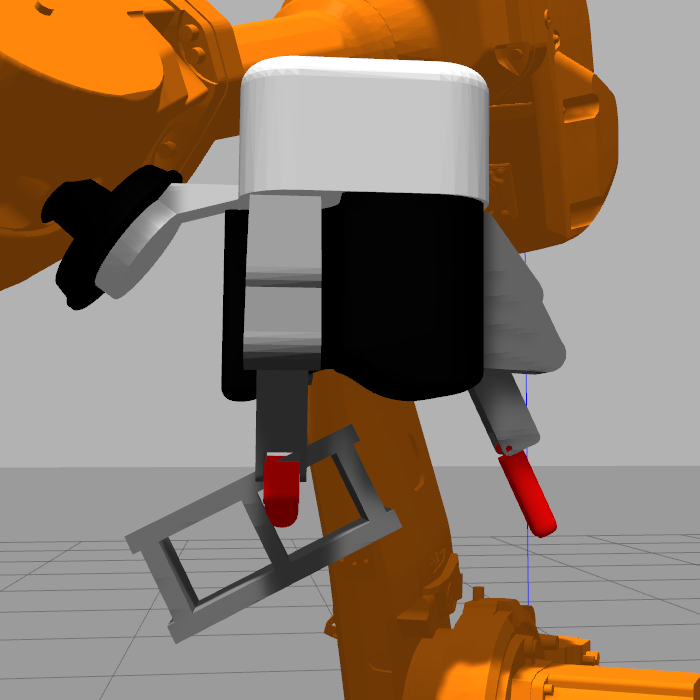} &
\includegraphics[width=0.19\linewidth, keepaspectratio]{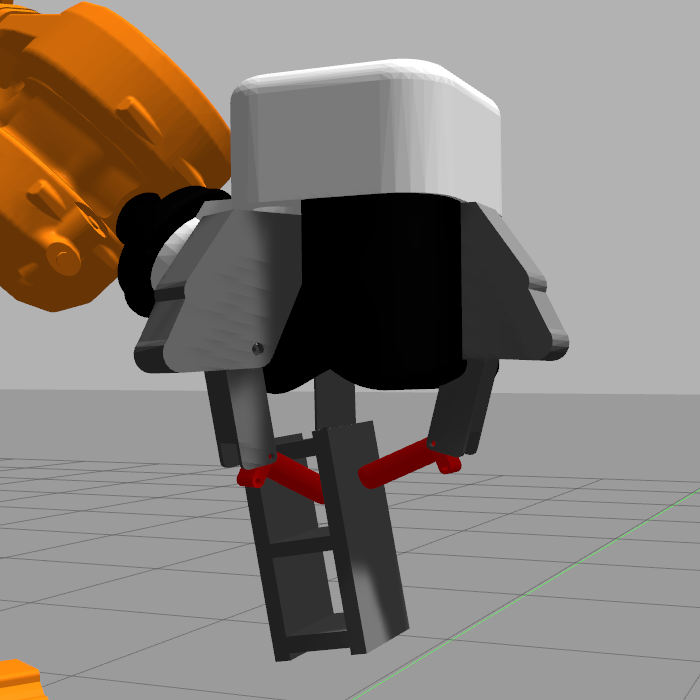} &
\includegraphics[width=0.19\linewidth, keepaspectratio]{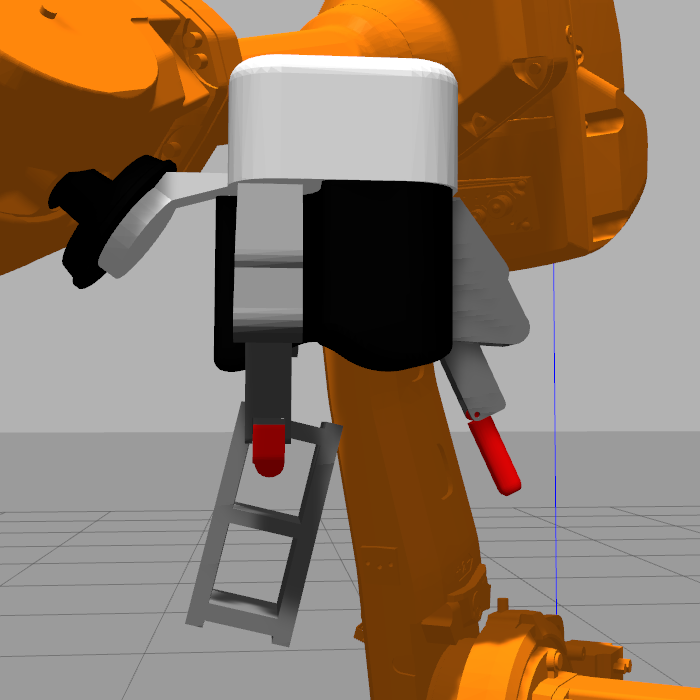}
\\
& \includegraphics[width=0.19\linewidth, keepaspectratio]{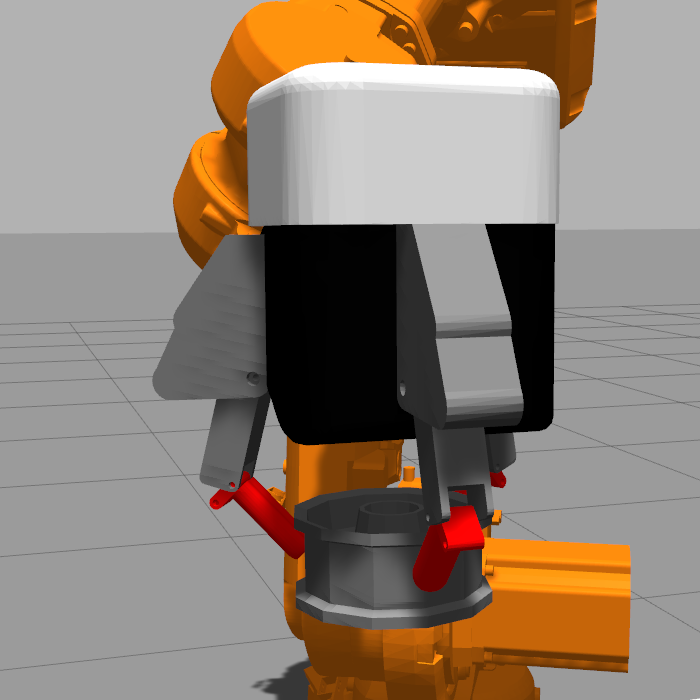} &
\includegraphics[width=0.19\linewidth, keepaspectratio]{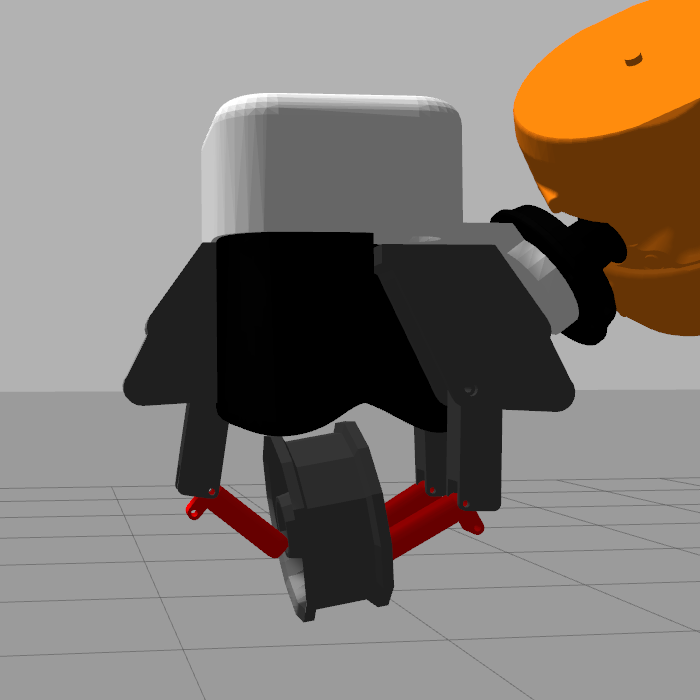} &
\includegraphics[width=0.19\linewidth, keepaspectratio]{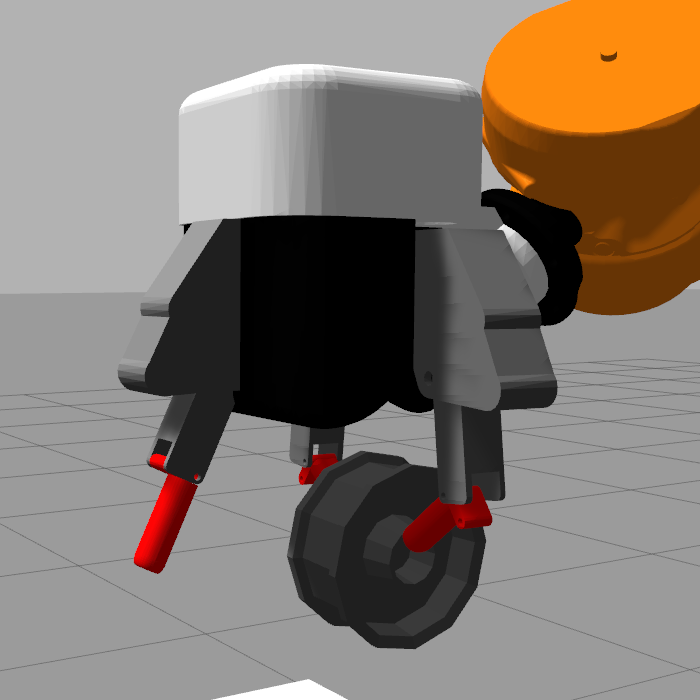} & 
\end{tabular}
\caption{Primitive grasp types for the three chosen objects. On the first row the grasp types for the bent pipe, on the second row for the cinder block, and on the third row for the pulley}
\label{fig:grasp_type}
\end{figure}

A set of primitive gripper configurations is determined for each of those objects for each of their stable position. These primitive gripper configurations can be sorted in different grasp types presented on \autoref{fig:grasp_type}. For each of these grasp type, several variants are created.

For each object is gathered the following number of primitive grasps:
\begin{samepage}
\begin{itemize}
\item bent pipe: 145 samples
\item cinder block: 141 samples
\item pulley: 118 samples
\end{itemize}
\end{samepage}

Around one hour is needed for a given object to register all those primitives.

\subsection{VAEs Training \& Quality Metric Computation}

\begin{figure*}
\centering
\medskip
\input{scheme_architecture}
\caption{HGG architecture. In blue the input layers, in green the hidden neural networks (NN) and in red the output layers. The hidden NN inner layers are fully connected layers, with hyperbolic tangent activation functions. The main encoding and main decoding NN have symmetrical inner architecture. The latent space dimensionality is one. The supplementary input for the tabletop equation allows to ensure that the generated grasp depends on it \cite{sohn_learning_2015}. This architecture is implemented with Tensorflow \cite{tensorflow2015-whitepaper} and Keras \cite{chollet2015keras} python libraries. The QGG has the same architecture, with an added output to the decoder for the grasp quality, with its associated output layer and hidden output NN. HGG and QGG have around 12000 trainable parameters each.}
\label{fig:archi}
\end{figure*}
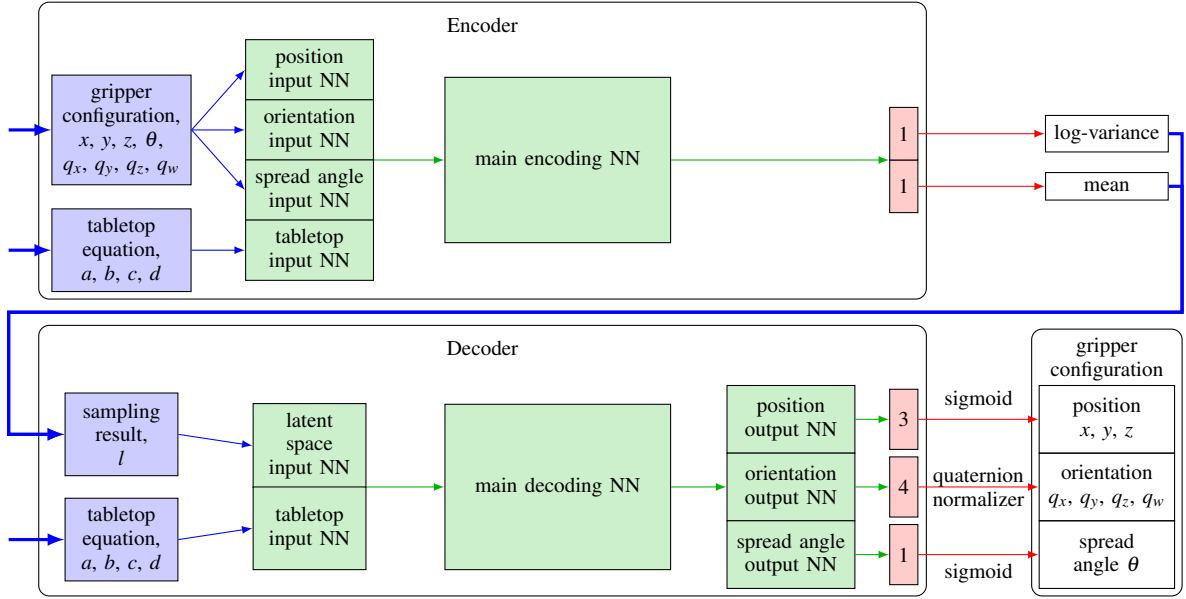

The architecture used for both HGG and QGG is displayed in \autoref{fig:archi}. 

before the training, the inputs and outputs data are normalized. This allows a faster training as the network does not have to scale its data by itself.

To make sure that the quaternion outputs by the decoder is a unit quaternion, a custom activation function is used to normalize the quaternion on the output layer of the decoder.

For each object, a HGG network is trained on the primitive grasp dataset, with the input-output architecture described in \autoref{fig:in_out}.

After the HGG training, 2000 grasps for each stable position of each object are generated by sampling in the HGG latent space and tested in simulation to compute their quality metric, in order to create the extended dataset for the QGG training. Some statistics about generated and primitive grasps quality are summarized on \autoref{fig:infos_dataset}.

The grasp quality mean and median of the primitive and generated grasps are close. This is expected as the VAE tries to reproduce the underlying distribution of the learning set. The goal is to explore the grasp space around the primitive grasps, and discover a collection of grasps with various quality. 

\begin{figure}[H]
\centering
\input{infos_dataset}
\caption{information summary about primitive and HGG generated grasps (step C of the workflow). The metric statistics are taking into account successful grasps only.}
\label{fig:infos_dataset}
\end{figure}

For Two objects, a global maximum better than the primitive grasps is found in the generated grasps. Indeed, the VAE learns to interpolate between the primitive grasps: in case a better grasp is "between" two primitive grasps, it is able to generate it. 

Regarding the bent pipe, no generated grasps are better than the best primitive grasp. Two possible explanations are:
\begin{itemize}
\item The global maximum may already be in the primitive dataset. It is not unlikely, as it is a human-crafted set of configuration, and humans tend to produce high quality grasps.
\item Some configurations of the dataset, among which primitives with best quality, are reconstructed poorly because of the trade-off between reconstruction and KL-divergence. The model may not generate configurations sufficiently close to those primitives to efficiently explore this part of the grasp space. This may be due to a too high compression, linked to a too small latent space.
\end{itemize}

Regarding the QGG network, it is trained for each object on the extended dataset constituted of the primitive grasps and the HGG generated grasps. Its performances are assessed on a grasp planning task in \autoref{sec:planning_test}.

\subsection{QGG Latent Space}

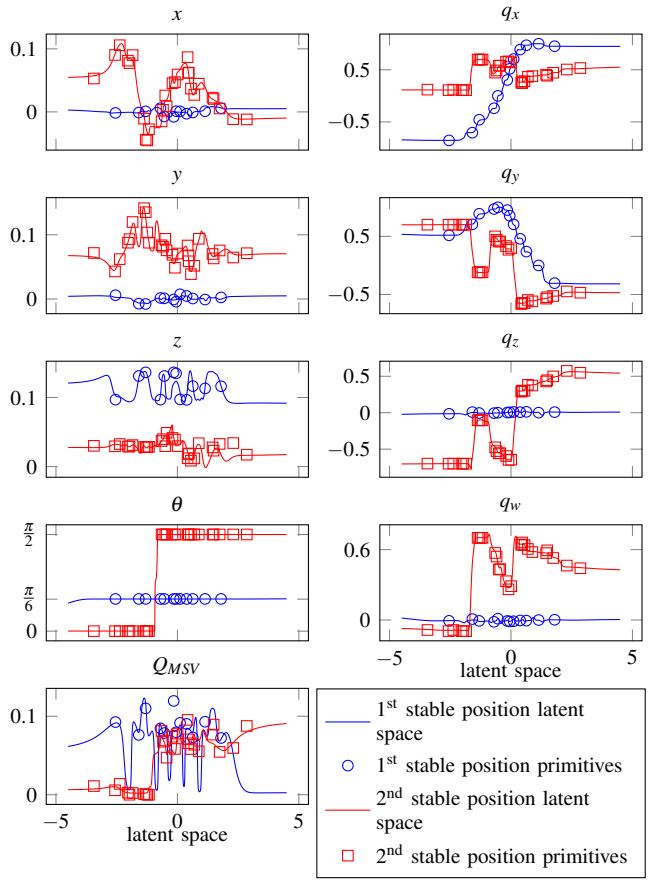
\begin{figure}[H]
\centering
\smallskip
\input{latent_space}
\caption{QGG latent space representation for the pulley object. The curves are obtained by sampling values in $\mathopen[-4.5, 4.5\mathclose]$ and passing them through the decoder. The outputs are gripper configurations stored in the latent space of the QGG. The scatter plots are obtained by passing some pulley primitive configurations through the encoder (only one third of the primitives displayed for readability).}
\label{fig:latent}
\end{figure}

The \autoref{fig:latent} shows how the QGG network has extracted the correlations between the gripper configuration parameters and the grasp quality value for the pulley object. The primitive grasps are evenly spread in the latent space, which is a direct consequence of the KL-divergence loss component. This is at the cost of a higher reconstruction error, but allows to sample in the latent space safely, without producing inconsistent configuration.

Due to contact points volatility in simulation, the computed metric has a variability and is not fully deterministic for a given configuration. Despite that, the QGG has successfully captured the overall tendency, without over-fitting on the noise. 

%% file: scheme_architecture.tex
\begin{tikzpicture}[every node/.append style={font=\footnotesize}]

\tikzstyle{inputNode}=[draw, fill=blue!20]
\tikzstyle{hiddenNode}=[draw, fill=black!30!green!20]
\tikzstyle{outputNode}=[draw, fill=red!20]
\tikzstyle{inputArrow}=[->, >=latex, blue]
\tikzstyle{hiddenArrow}=[->, >=latex, black!30!green]
\tikzstyle{outputArrow}=[->, >=latex, red]

%encoder
\draw[rounded corners] (-0.6,2.2) rectangle (11.2,-1.75);
\node[] at (5.3,1.9) {Encoder};
\node[inputNode, minimum height=1.1cm, minimum width=1.85cm, align=center] (conf-enc) at (0.5, 0.5){gripper \\ configuration, \\ $x$, $y$, $z$, $\theta$, \\ $q_x$, $q_y$, $q_z$, $q_w$};
\node[inputNode, minimum height=1.1cm, minimum width=1.85cm, align=center] (table-enc) at (0.5, -1.1) {tabletop \\ equation, \\ $a$, $b$, $c$, $d$};

%preprocessing
\node[minimum height=3.2cm, minimum width=1.7cm](dummy-node-preenc) at (3., 0.1){};
\node[hiddenNode, minimum height=0.8cm, minimum width=1.7cm, align=center](prepos-enc) at (3.,1.3) {position \\ input NN};
\node[hiddenNode, minimum height=0.8cm, minimum width=1.7cm, align=center](preori-enc) at (3.,0.5) {orientation \\ input NN};
\node[hiddenNode, minimum height=0.8cm, minimum width=1.7cm, align=center](presp-enc) at (3.,-0.3) {spread angle \\ input NN};
\node[hiddenNode, minimum height=0.8cm, minimum width=1.7cm, align=center](pretable-enc) at (3.,-1.1) {tabletop \\ input NN};

% main encoding
\node[hiddenNode, minimum height=2.2cm, minimum width=3cm] (main-enc) at (6.3,0.1) {main encoding NN};

\node[minimum height=1.4cm, minimum width=0.375cm](dummy-node-output) at (10.9, 0.1){};
\node[outputNode, minimum height=0.7cm] (out1-enc) at (10.9,-0.25) {1};
\node[outputNode, minimum height=0.7cm] (out2-enc) at (10.9,0.45) {1};

\node[draw, minimum width=1.63cm] (mean) at (13.6,-0.25) {mean};
\node[draw, minimum width=1.63cm] (variance) at (13.6,0.45) {log-variance};

%decoder
\draw[rounded corners] (-0.6,-2.1) rectangle (11.2,-5.7);
\node[font=\footnotesize] at (5.3,-2.4) {Decoder};
\node[inputNode, minimum width=1.5cm, minimum height=1.1cm, align=center] (in1-dec) at (0.5,-4.95) {tabletop \\ equation, \\ $a$, $b$, $c$, $d$};
\node[inputNode, minimum width=1.5cm, minimum height=1.1cm, align=center] (in2-dec) at (0.5,-3.55) {sampling \\ result, \\ $l$};

%preprocessing
\node[minimum height=2.2cm, minimum width=1.5cm](dummy-node-predec) at (3., -4.25){};
\node[hiddenNode, minimum height=1.1cm, minimum width=1.5cm, align=center](prelat-dec) at (3.,-3.7) {latent \\ space \\ input NN};
\node[hiddenNode, minimum height=1.1cm, minimum width=1.5cm, align=center](pretable-dec) at (3.,-4.8) {tabletop \\ input NN};

% main decoding
\node[hiddenNode, minimum height=2.2cm, minimum width=3cm] (main-dec) at (6.3,-4.25) {main decoding NN};

% postprocessing
\node[hiddenNode, minimum height=0.9cm, minimum width=1.7cm, align=center](postpos-dec) at (9.4,-3.35) {position \\ output NN};
\node[hiddenNode, minimum height=0.9cm, minimum width=1.7cm, align=center](postori-dec) at (9.4,-4.25) {orientation \\ output NN};
\node[hiddenNode, minimum height=0.9cm, minimum width=1.7cm, align=center](postsp-dec) at (9.4,-5.15) {spread angle \\ output NN};

\node[outputNode, minimum height=0.8cm] (outpos-dec) at (10.9,-3.35) {3};
\node[outputNode, minimum height=0.8cm] (outori-dec) at (10.9,-4.25) {4};
\node[outputNode, minimum height=0.8cm] (outspread-dec) at (10.9,-5.15) {1};

\draw[rounded corners] (12.6, -2.15) rectangle (14.6, -5.70);
\node[align=center] (config) at (13.6,-2.55) {gripper \\ configuration};
\node[draw, align=center, minimum height=0.9cm, minimum width=1.8cm] (pos) at (13.6,-3.35) {position \\ $x$, $y$, $z$};
\node[draw, align=center, minimum height=0.9cm, minimum width=1.8cm] (ori) at (13.6,-4.25) {orientation \\ $q_x$, $q_y$, $q_z$, $q_w$};
\node[draw, align=center, minimum height=0.9cm, minimum width=1.8cm] (spread) at (13.6,-5.15) {spread \\ angle $\theta$};

% arrows encoder
\draw[inputArrow, very thick] (-1, 0.5) -- (conf-enc.west);
\draw[inputArrow, very thick] (-1, -1.1) -- (table-enc.west);
\draw[inputArrow] (conf-enc.east) -- (prepos-enc.west);
\draw[inputArrow] (conf-enc.east) -- (preori-enc.west);
\draw[inputArrow] (conf-enc.east) -- (presp-enc.west);
\draw[inputArrow] (table-enc.east) -- (pretable-enc.west);
%\draw[hiddenArrow] (prepos-enc.east) -- (main-enc.west);
%\draw[hiddenArrow] (preori-enc.east) -- (main-enc.west);
%\draw[hiddenArrow] (presp-enc.east) -- (main-enc.west);
%\draw[hiddenArrow] (pretable-enc.east) -- (main-enc.west);
\draw[hiddenArrow] (dummy-node-preenc.east) -- (main-enc.west);
\draw[hiddenArrow] (main-enc.east) -- (dummy-node-output.west);
\draw[outputArrow] (out1-enc) -- (mean);
\draw[outputArrow] (out2-enc) -- (variance);

% arrows decoder
\draw[inputArrow, very thick] (-1, -4.95) -- (in1-dec);
\draw[inputArrow, very thick] (variance.east) -| (14.6, -0.7) |- (-1, -1.925) -| (-1,-3) |- (in2-dec.west);
\draw[inputArrow, very thick] (mean.east) -| (14.6, -0.7) |- (-1, -1.925) -| (-1,-3) |- (in2-dec.west);
\draw[inputArrow] (in2-dec.east) -- (prelat-dec.west);
\draw[inputArrow] (in1-dec.east) -- (pretable-dec.west);
%\draw[hiddenArrow] (prelat-dec.east) -- (main-dec.west);
%\draw[hiddenArrow] (pretable-dec.east) -- (main-dec.west);
\draw[hiddenArrow] (dummy-node-predec.east) -- (main-dec.west);
%\draw[hiddenArrow] (main-dec.east) -- (postpos-dec.west);
\draw[hiddenArrow] (main-dec.east) -- (postori-dec.west);
%\draw[hiddenArrow] (main-dec.east) -- (postsp-dec.west);
\draw[hiddenArrow] (postpos-dec.east) -- (outpos-dec.west);
\draw[hiddenArrow] (postori-dec.east) -- (outori-dec.west);
\draw[hiddenArrow] (postsp-dec.east) -- (outspread-dec.west);

\draw[outputArrow] (outpos-dec.east) -- (pos.west) node[midway, above, black, align=center]{sigmoid};
\draw[outputArrow] (outori-dec.east) -- (ori.west) node[midway, black, align=center]{quaternion \\ normalizer};
\draw[outputArrow] (outspread-dec.east) -- (spread.west) node[midway, below, black, align=center]{sigmoid};
\end{tikzpicture}

%% file: infos_dataset.tex
\footnotesize
\begin{tabular}{l l l c  c }
& \multicolumn{2}{c}{} & \textbf{generated set} & \textbf{primitive set} \\
\toprule
\multirow{2}{1.2cm}[-27pt]{\textbf{bent pipe}} & \multicolumn{2}{>{\arraybackslash}m{2.6cm}}{total number of grasps} & 4000 & 145 \\
\cmidrule{2-5}
& \multicolumn{2}{>{\arraybackslash}m{2.6cm}}{number of successful grasps} & 2845 & 141 \\
\cmidrule{2-5}
& \multirow{3}{1cm}[-5pt]{metric statistics} & median & 0.1022 & 0.1018 \\
\cmidrule{3-5}
& & mean & 0.1035 & 0.1047 \\
\cmidrule{3-5}
& & maximum & 0.2128 & 0.2257 \\
\midrule
\midrule
\multirow{2}{1.2cm}[-27pt]{\textbf{cinder block}} & \multicolumn{2}{>{\arraybackslash}m{2.6cm}}{total number of grasps} & 6000 & 141 \\
\cmidrule{2-5}
& \multicolumn{2}{>{\arraybackslash}m{2.6cm}}{number of successful grasps} & 5690 & 141 \\
\cmidrule{2-5}
& \multirow{3}{1cm}[-5pt]{metric statistics} & median & 0.0675 & 0.0670 \\
\cmidrule{3-5}
& & mean & 0.0535 & 0.0563 \\
\cmidrule{3-5}
& & maximum & 0.1877 & 0.1041 \\
\midrule
\midrule
\multirow{2}{1.2cm}[-27pt]{\textbf{pulley}} & \multicolumn{2}{>{\arraybackslash}m{2.6cm}}{total number of grasps} & 4000 & 118 \\
\cmidrule{2-5}
& \multicolumn{2}{>{\arraybackslash}m{2.6cm}}{number of successful grasps} & 3591 & 111 \\
\cmidrule{2-5}
& \multirow{3}{1cm}[-5pt]{metric statistics} & median & 0.0739 & 0.0730 \\
\cmidrule{3-5}
& & mean & 0.0653 & 0.0648 \\
\cmidrule{3-5}
& & maximum & 0.3115 & 0.1195 \\
\bottomrule
\end{tabular}

%% file: latent_space.tex
\begin{tikzpicture}
\pgfplotsset{every axis/.append style={label style={font=\footnotesize}, tick label style={font=\footnotesize}, title style={font=\footnotesize}}}
\pgfplotsset{title style={at={(0.5,0.90)}}}
\pgfplotsset{xlabel style={at={(0.5,-0.11)}}}
\pgfplotsset{ylabel style={at={(-0.15,0.5)}}}
\begin{groupplot}[group style={group size=2 by 5, x descriptions at=edge bottom, vertical sep=1.8em,horizontal sep=6ex}, height=0.13\textheight,width=0.285\textwidth, xlabel=latent space]
	\nextgroupplot[title=$x$, ytick={0, 0.1}, yticklabels={0, 0.1}]
	\addplot[blue] table [x=latent_value, y=Hand_X, col sep=comma] 		{latent_space_0.csv};
	\addplot[blue, only marks, mark=o] table [x=latent_value, y=Hand_X, col sep=comma] {primitive_0.csv};
	\addplot[red] table [x=latent_value, y=Hand_X, col sep=comma] {latent_space_1.csv};
	\addplot[red, only marks, mark=square] table [x=latent_value, y=Hand_X, col sep=comma] {primitive_1.csv};
	
	\nextgroupplot[title=$q_x$, ytick={-0.5, 0.5}, yticklabels={$-0.5$, 0.5}]
	\addplot[blue] table [x=latent_value, y=Hand_Q_X, col sep=comma] {latent_space_0.csv};
	\addplot[blue, only marks, mark=o] table [x=latent_value, y=Hand_Q_X, col sep=comma] {primitive_0.csv};
	\addplot[red] table [x=latent_value, y=Hand_Q_X, col sep=comma] {latent_space_1.csv};
	\addplot[red, only marks, mark=square] table [x=latent_value, y=Hand_Q_X, col sep=comma] {primitive_1.csv};
	
	\nextgroupplot[title=$y$, ytick={0, 0.1}, yticklabels={0, 0.1}]
	\addplot[blue] table [x=latent_value, y=Hand_Y, col sep=comma] {latent_space_0.csv};
	\addplot[blue, only marks, mark=o] table [x=latent_value, y=Hand_Y, col sep=comma] {primitive_0.csv};
	\addplot[red] table [x=latent_value, y=Hand_Y, col sep=comma] {latent_space_1.csv};
	\addplot[red, only marks, mark=square] table [x=latent_value, y=Hand_Y, col sep=comma] {primitive_1.csv};
	
	\nextgroupplot[title=$q_y$, ytick={-0.5, 0.5}, yticklabels={$-0.5$, 0.5}]
	\addplot[blue] table [x=latent_value, y=Hand_Q_Y, col sep=comma] {latent_space_0.csv};
	\addplot[blue, only marks, mark=o] table [x=latent_value, y=Hand_Q_Y, col sep=comma] {primitive_0.csv};
	\addplot[red] table [x=latent_value, y=Hand_Q_Y, col sep=comma] {latent_space_1.csv};
	\addplot[red, only marks, mark=square] table [x=latent_value, y=Hand_Q_Y, col sep=comma] {primitive_1.csv};
	
	\nextgroupplot[title=$z$, ytick={0, 0.1}, yticklabels={0, 0.1}]
	\addplot[blue] table [x=latent_value, y=Hand_Z, col sep=comma] {latent_space_0.csv};
	\addplot[blue, only marks, mark=o] table [x=latent_value, y=Hand_Z, col sep=comma] {primitive_0.csv};
	\addplot[red] table [x=latent_value, y=Hand_Z, col sep=comma] {latent_space_1.csv};
	\addplot[red, only marks, mark=square] table [x=latent_value, y=Hand_Z, col sep=comma] {primitive_1.csv};
			
	\nextgroupplot[title=$q_z$]
	\addplot[blue] table [x=latent_value, y=Hand_Q_Z, col sep=comma] {latent_space_0.csv};
	\addplot[blue, only marks, mark=o] table [x=latent_value, y=Hand_Q_Z, col sep=comma] {primitive_0.csv};
	\addplot[red] table [x=latent_value, y=Hand_Q_Z, col sep=comma] {latent_space_1.csv};
	\addplot[red, only marks, mark=square] table [x=latent_value, y=Hand_Q_Z, col sep=comma] {primitive_1.csv};
		
	\nextgroupplot[title=$\theta$, ytick={0, 0.52, 1.57}, yticklabels={0, $\frac{\pi}{6}$, $\frac{\pi}{2}$}]
	\addplot[blue] table [x=latent_value, y=Hand_Abd, col sep=comma] {latent_space_0.csv};
	\addplot[blue, only marks, mark=o] table [x=latent_value, y=Hand_Abd, col sep=comma] {primitive_0.csv};
	\addplot[red] table [x=latent_value, y=Hand_Abd, col sep=comma] {latent_space_1.csv};
	\addplot[red, only marks, mark=square] table [x=latent_value, y=Hand_Abd, col sep=comma] {primitive_1.csv};
	
	\nextgroupplot[title=$q_w$, ytick={0, 0.6}, yticklabels={0, 0.6}, xlabel=latent space, xtick={-5, 0, 5}, xticklabels={$-5$, 0, 5}, ]
	\addplot[blue] table [x=latent_value, y=Hand_Q_W, col sep=comma] {latent_space_0.csv};
	\addplot[blue, only marks, mark=o] table [x=latent_value, y=Hand_Q_W, col sep=comma] {primitive_0.csv};
	\addplot[red] table [x=latent_value, y=Hand_Q_W, col sep=comma] {latent_space_1.csv};
	\addplot[red, only marks, mark=square] table [x=latent_value, y=Hand_Q_W, col sep=comma] {primitive_1.csv};
	
	\nextgroupplot[title=$Q_{MSV}$, ytick={0, 0.1}, yticklabels={0, 0.1}, legend style={legend cell align=left, legend pos=outer north east, font=\footnotesize, nodes={text width=3.4cm, text depth={}}}]
	\addplot[blue] table [x=latent_value, y=Grasp_Quality, col sep=comma] {latent_space_0.csv};
	\addplot[blue, only marks, mark=o] table [x=latent_value, y=Grasp_Quality, col sep=comma] {primitive_0.csv};
	\addplot[red] table [x=latent_value, y=Grasp_Quality, col sep=comma] {latent_space_1.csv};
	\addplot[red, only marks, mark=square] table [x=latent_value, y=Grasp_Quality, col sep=comma] {primitive_1.csv};
	\legend{1\textsuperscript{st} stable position latent space, 1\textsuperscript{st} stable position primitives, 2\textsuperscript{nd} stable position latent space, 2\textsuperscript{nd} stable position primitives}
\end{groupplot}
\end{tikzpicture}

%% file: planning_trial.tex
To validate the grasp space exploration workflow, grasp planning trials are conducted on each object. The grasp planning algorithm is described in \autoref{alg:trial}.

\begin{algorithm}[b]
\centering
\footnotesize
\input{plan_trial_algo}
\caption{\footnotesize Grasp planning algorithm.}
\label{alg:trial}
\end{algorithm}

This planning procedure is executed on 1000 distinct object poses for each stable position of each object. The position of the object frame projection on the tabletop plane is chosen randomly inside a $10 \times 10$ centimeters square, and its orientation relative to the vertical axis is also drawn randomly between 0 and $2 \pi$. 

Three parameters are monitored to assess the performances of the presented workflow:
\begin{enumerate}
\item the grasp success rate.
\item the number of collision and reachability checking iterations needed to find three admissible grasps (\autoref{alg:trial} line 5), as it is the most time consuming step. Indeed, the presented workflow is object-centric. It does not take into account the arm kinematic and environment, so depending on the object pose in the robot workspace, the probability of sampling an admissible configurations in the QGG latent space vary.
\item the grasp quality relative prediction error.
\end{enumerate}
These parameters are shown on \autoref{fig:perf_recap}.

\begin{figure}[ht]
\centering
\input{perf_recap}
\caption{Performances on grasp planning trials.}
\label{fig:perf_recap}
\end{figure}

The mean relative prediction error has the same order of magnitude than the computed metric noise.

The low number of collision and reachability checking iterations shows that all grasp types and their variants are evenly distributed in the latent space. Indeed, some grasp types or variants within a grasp type are reachable only for some object poses relative to the robot.

The low failure rate shows that the procedure presented in this work successfully explore and reproduce the grasp space, as it is able to generate reliably high quality grasps for various object poses.

%% file: plan_trial_algo.tex
\begin{algorithmic}[1]
\STATE \emph{grasp candidate list} $\gets \emptyset$
\WHILE{length(\emph{grasp candidate list}) $< 3$}
\STATE \emph{configuration} $ \gets$ QGG decoding of a sampled value in its latent space
\IF {\emph{configuration} predicted grasp quality $>$ threshold}
\IF {\emph{configuration} is collision free \AND kinematically reachable}
\STATE append \emph{configuration} to \emph{grasp candidate list}
\ENDIF
\ENDIF
\ENDWHILE
\STATE execute the grasp with highest predicted grasp quality among \emph{grasp candidate list}
\end{algorithmic}

%% file: perf_recap.tex
\footnotesize
\begin{tabular}{l >{\centering\arraybackslash}m{1.8cm} >{\centering\arraybackslash}m{1.8cm} >{\centering\arraybackslash}m{1.8cm}}
& \textbf{1) success rate (\%)} & \textbf{2) \autoref{alg:trial} line 5 mean iterations} & \textbf{3) mean quality prediction error (\%)} \\
\toprule
bent pipe & 100 & 7.6 & 15.6 \\
cinder block & 100 & 5.8 & 7.5 \\
pulley & 99.7 & 7.4 & 14.8 \\
\bottomrule

\end{tabular}

%% file: conclusion.tex
This work presents an efficient method for grasp space exploration. It explores a high dimensional grasp space by focusing the search around human inputs, and take into account analytic grasp quality criterion. This procedure was then used to successfully plan grasps in simulation.

Various tracks can be investigated in future works. First, the effect of a latent space of higher dimension on the grasp space exploration needs to be assessed. Indeed, using a larger latent space may improve the exhaustiveness of the exploration by reducing information loss due to compression. Moreover, a reduction of the number of human inputs required per object would be useful to scale this method to several objects. Furthermore, in this study, the grasp planning performances were assessed in simulation only. These results need to be confirmed on a real setup. Finally, this space exploration procedure could be used to constitute a grasp dataset with high quality grasps to be learned by a data-driven grasp planner. Indeed, this may allow to generalize to unseen objects.

%% file: root.bbl
\begin{thebibliography}{10}
\providecommand{\url}[1]{#1}
\csname url@rmstyle\endcsname
\providecommand{\newblock}{\relax}
\providecommand{\bibinfo}[2]{#2}
\providecommand\BIBentrySTDinterwordspacing{\spaceskip=0pt\relax}
\providecommand\BIBentryALTinterwordstretchfactor{4}
\providecommand\BIBentryALTinterwordspacing{\spaceskip=\fontdimen2\font plus
\BIBentryALTinterwordstretchfactor\fontdimen3\font minus
  \fontdimen4\font\relax}
\providecommand\BIBforeignlanguage[2]{{%
\expandafter\ifx\csname l@#1\endcsname\relax
\typeout{** WARNING: IEEEtran.bst: No hyphenation pattern has been}%
\typeout{** loaded for the language `#1'. Using the pattern for}%
\typeout{** the default language instead.}%
\else
\language=\csname l@#1\endcsname
\fi
#2}}

\bibitem{townsend_barretthand_2000}
W.~Townsend, ``The {BarrettHand} grasper – programmably flexible part
  handling and assembly,'' \emph{Industrial Robot}, vol.~27, no.~3, pp.
  181--188, 2000.

\bibitem{sahbani_overview_2012}
A.~Sahbani, S.~El-Khoury, and P.~Bidaud, ``An overview of 3d object grasp
  synthesis algorithms,'' \emph{Robotics and Autonomous Systems}, vol.~60,
  no.~3, pp. 326--336, 2012.

\bibitem{berenson_grasp_2007}
D.~{Berenson}, R.~{Diankov}, {Koichi Nishiwaki}, {Satoshi Kagami}, and
  J.~{Kuffner}, ``Grasp planning in complex scenes,'' in \emph{IEEE-RAS
  International Conference on Humanoid Robots}, 2007, pp. 42--48.

\bibitem{roa_grasp_2008}
M.~A. {Roa}, R.~{Suarez}, and J.~{Rosell}, ``Grasp space generation using
  sampling and computation of independent regions,'' in \emph{IEEE/RSJ
  International Conference on Intelligent Robots and Systems}, 2008, pp.
  2258--2263.

\bibitem{xue_grasp_2007}
Z.~{Xue}, J.~M. {Zoellner}, and R.~{Dillmann}, ``Grasp planning: Find the
  contact points,'' in \emph{IEEE International Conference on Robotics and
  Biomimetics (ROBIO)}, 2007, pp. 835--840.

\bibitem{zhao_grasp_2020}
Z.~Zhao, W.~Shang, H.~He, and Z.~Li, ``Grasp prediction and evaluation of
  multi-fingered dexterous hands using deep learning,'' \emph{Robotics and
  Autonomous Systems}, vol. 129, p. 103550, 2020.

\bibitem{pinto_supersizing_2015}
L.~Pinto and A.~Gupta, ``Supersizing self-supervision: Learning to grasp from
  50k tries and 700 robot hours,'' in \emph{IEEE International Conference on
  Robotics and Automation (ICRA)}, 2016, pp. 3406--3413.

\bibitem{depierre_jacquard:_2018}
A.~Depierre, E.~Dellandréa, and L.~Chen, ``Jacquard: A large scale dataset for
  robotic grasp detection,'' in \emph{{IEEE}/{RSJ} International Conference on
  Intelligent Robots and Systems ({IROS})}, 2018, pp. 3511--3516.

\bibitem{levine_learning_2018}
S.~Levine, P.~Pastor, A.~Krizhevsky, J.~Ibarz, and D.~Quillen, ``Learning
  hand-eye coordination for robotic grasping with deep learning and large-scale
  data collection,'' \emph{The International Journal of Robotics Research},
  vol.~37, no.~4, pp. 421--436, 2018.

\bibitem{mahler_dex-net_2017}
J.~Mahler, J.~Liang, S.~Niyaz, M.~Laskey, R.~Doan, X.~Liu, J.~A. Ojea, and
  K.~Goldberg, ``Dex-net 2.0: Deep learning to plan robust grasps with
  synthetic point clouds and analytic grasp metrics,'' in \emph{Robotics:
  Science and Systems (RSS)}, 2017.

\bibitem{Riedlinger_Model_2020}
M.~A. c.~{Riedlinger}, M.~{Voelk}, K.~{Kleeberger}, M.~U. {Khalid}, and
  R.~{Bormann}, ``Model-free grasp learning framework based on physical
  simulation,'' in \emph{ISR 2020; 52th International Symposium on Robotics},
  2020, pp. 1--8.

\bibitem{Mousavian_graspnet_2019}
A.~{Mousavian}, C.~{Eppner}, and D.~{Fox}, ``6-dof graspnet: Variational grasp
  generation for object manipulation,'' in \emph{2019 IEEE/CVF International
  Conference on Computer Vision (ICCV)}, 2019, pp. 2901--2910.

\bibitem{santina_learning_2019}
C.~D. Santina, V.~Arapi, G.~Averta, F.~Damiani, G.~Fiore, A.~Settimi, M.~G.
  Catalano, D.~Bacciu, A.~Bicchi, and M.~Bianchi, ``Learning from humans how to
  grasp: A data-driven architecture for autonomous grasping with
  anthropomorphic soft hands,'' \emph{{IEEE} Robotics and Automation Letters},
  vol.~4, no.~2, pp. 1533--1540, 2019.

\bibitem{choi_learning_2018}
C.~Choi, W.~Schwarting, J.~{DelPreto}, and D.~Rus, ``Learning object grasping
  for soft robot hands,'' \emph{{IEEE} Robotics and Automation Letters},
  vol.~3, no.~3, pp. 2370--2377, 2018.

\bibitem{koenig_design_2004}
N.~Koenig and A.~Howard, ``Design and use paradigms for gazebo, an open-source
  multi-robot simulator,'' in \emph{{IEEE}/{RSJ} International Conference on
  Intelligent Robots and Systems ({IROS})}, vol.~3, 2004, pp. 2149--2154.

\bibitem{drost_model_2010}
B.~{Drost}, M.~{Ulrich}, N.~{Navab}, and S.~{Ilic}, ``Model globally, match
  locally: Efficient and robust 3d object recognition,'' in \emph{IEEE Computer
  Society Conference on Computer Vision and Pattern Recognition}, 2010, pp.
  998--1005.

\bibitem{roa_grasp_2015}
M.~A. Roa and R.~Suárez, ``Grasp quality measures: review and performance,''
  \emph{Autonomous Robots}, vol.~38, no.~1, pp. 65--88, 2015.

\bibitem{murray_mathematical_1994}
R.~M. Murray, Z.~Li, and S.~S. Sastry, \emph{A Mathematical Introduction to
  Robotic Manipulation}.\hskip 1em plus 0.5em minus 0.4em\relax {CRC} Press,
  1994.

\bibitem{prattichizzo_grasping_2008}
D.~Prattichizzo and J.~C. Trinkle, ``Grasping,'' in \emph{Handbook of
  Robotics}.\hskip 1em plus 0.5em minus 0.4em\relax Springer, 2008, pp.
  671--700.

\bibitem{kingma_auto_2014}
D.~Kingma and M.~Welling, ``Auto-encoding variational bayes,'' in
  \emph{International Conference on Learning Representations (ICLR)}, 2014.

\bibitem{sohn_learning_2015}
K.~Sohn, X.~Yan, and H.~Lee, ``Learning structured output representation using
  deep conditional generative models,'' in \emph{International Conference on
  Neural Information Processing Systems}, ser. NIPS'15, vol.~2.\hskip 1em plus
  0.5em minus 0.4em\relax Cambridge, MA, USA: MIT Press, 2015, pp. 3483--3491.

\bibitem{tensorflow2015-whitepaper}
M.~Abadi \emph{et~al.}, ``{TensorFlow}: Large-scale machine learning on
  heterogeneous systems,'' 2015, software available from
  \url{www.tensorflow.org}.

\bibitem{chollet2015keras}
F.~Chollet \emph{et~al.}, ``Keras,'' 2015, software available from
  \url{www.keras.io}.

\end{thebibliography}
